\documentclass[journal]{IEEEtran}
\usepackage{amsmath,graphicx}
\usepackage{booktabs}
\usepackage{cite}
\usepackage{multirow}
\usepackage{multicol}
\usepackage{tabularx}
\usepackage{float}
\usepackage{makecell}
\usepackage{subfigure}
\usepackage{enumitem}
\usepackage{amsfonts,amssymb}
\usepackage[colorlinks]{hyperref}
\usepackage{cleveref}
\usepackage{bm}
\usepackage{graphicx}
\usepackage{color}
\usepackage{soul}

\usepackage{bigstrut} 
\usepackage{makecell} 

\crefname{figure}{Fig.}{Figs.}
\Crefname{figure}{Figure}{Figures}
\crefname{section}{Sect.}{Sects.}
\Crefname{section}{Section}{Section}
\crefname{equation}{Eq.}{Eqs.}
\Crefname{equation}{Equation}{Equation}
\crefname{table}{Table}{Tables.}
\Crefname{table}{Table}{Tables}

\ifCLASSINFOpdf

\else

\fi

\begin{document}

\title{Generic Knowledge Boosted Pre-training For Remote Sensing Images}

\author{Ziyue~Huang, Mingming Zhang, Yuan Gong, Qingjie~Liu, ~\IEEEmembership{Member,~IEEE,}, and Yunhong~Wang,~\IEEEmembership{Fellow,~IEEE}%

\thanks{This work was supported by the National Natural Science Foundation of China	under Grant 62176017. \textit{(Corresponding author: Qingjie Liu)}}
\thanks{
Ziyue Huang, Mingming Zhang, Qingjie Liu, and Yunhong Wang are with the State Key Laboratory of Virtual Reality Technology and Systems, Beihang University, Beijing 100191, China, and also with the Hangzhou Innovation Institute, Beihang University, Hangzhou 310051, China (e-mail: ziyuehuang@buaa.edu.cn; sara\_@buaa.edu.cn; qingjie.liu@buaa.edu.cn; yhwang@buaa.edu.cn).}
\thanks{Yuan Gong are with School of Software and Microelectronics, Peking University, Beijing 100871, China. (e-mail: alangy@stu.pku.edu.cn). }
}

\markboth{IEEE Transactions on Geoscience and Remote Sensing}%
{Shell \MakeLowercase{\textit{et al.}}: Bare Demo of IEEEtran.cls for IEEE Journals}

\maketitle

\begin{abstract}
Deep learning models are essential for scene classification, change detection, land cover segmentation, and other remote sensing image understanding tasks. 
Most backbones of existing remote sensing deep learning models are typically initialized by pre-trained weights obtained from ImageNet pre-training (IMP). 
However, domain gaps exist between remote sensing images and natural images (\textit{e.g.}, ImageNet), making deep learning models initialized by pre-trained weights of IMP perform poorly for remote sensing image understanding. 
Although some pre-training methods are studied in the remote sensing community, current remote sensing pre-training methods face the problem of vague generalization by only using remote sensing images. 
In this paper, we propose a novel remote sensing pre-training framework, Generic Knowledge Boosted Remote Sensing Pre-training (GeRSP), to learn robust representations from remote sensing and natural images for remote sensing understanding tasks. 
GeRSP contains two pre-training branches: (1) A self-supervised pre-training branch is adopted to learn domain-related representations from unlabeled remote sensing images. 
(2) A supervised pre-training branch is integrated into GeRSP for general knowledge learning from labeled natural images. 
Moreover, GeRSP combines two pre-training branches using a teacher-student architecture to simultaneously learn representations with general and special knowledge, which generates a powerful pre-trained model for deep learning model initialization. 
Finally, we evaluate GeRSP and other remote sensing pre-training methods on three downstream tasks, \textit{i.e.}, object detection, semantic segmentation, and scene classification. 
The extensive experimental results consistently demonstrate that GeRSP can effectively learn robust representations in a unified manner, improving the performance of remote sensing downstream tasks. 
Code and pre-trained models: https://github.com/floatingstarZ/GeRSP.
\end{abstract}

\begin{IEEEkeywords}
Remote sensing image, pre-training, self-supervised learning
\end{IEEEkeywords}

%
\IEEEpeerreviewmaketitle

\section{Introduction}
\IEEEPARstart{D}{eep} learning models have been widely used in remote sensing (RS) image understanding tasks, such as detection \cite{RoITrans}, segmentation \cite{marmanis2016semantic}, and scene classification \cite{cheng2017remote}. 
Most of these interpretation models are initialized with ImageNet \cite{ImageNet} pre-trained weights. 
Although it has been proved that ImageNet pre-trained models generalize well to the RS interpretation tasks, domain gaps still exist between RS and natural images due to different capture views, image resolutions, and object appearances, which impedes the RS image understanding performance.
This puts forward an urgent requirement for Remote Sensing Pre-training (RSP) techniques \cite{Wang_RS_Pretrain}. 

\begin{figure}[!tb]
    \centering
    \includegraphics[width=1.0\linewidth]{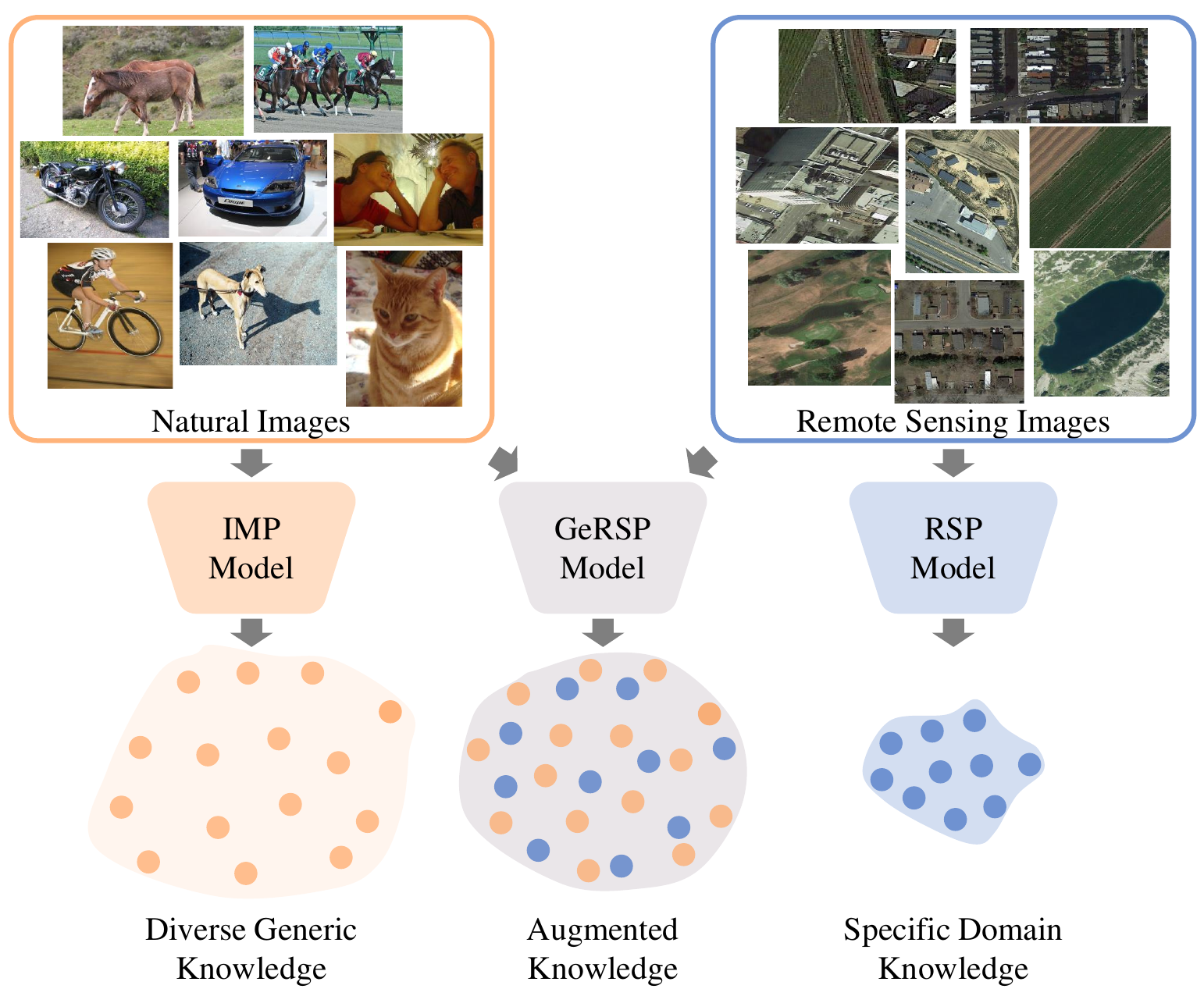}
	\caption{
 RS images encompass a wealth of domain-specific knowledge, whereas natural images offer a broader range of diverse generic image knowledge. 
 The motivation of the GeRSP is to enhance the generalization performance of RSP by leveraging the diversity present in natural images.}
	\label{fig:GeRSP_motivation}
\end{figure}

Most RSP methods draw inspiration from general pre-training methodologies, such as MoCo \cite{MoCo}, SimCLR\cite{SimCLR}, and MAE \cite{MAE}, which can be categorized as supervised and self-supervised paradigms. 
The supervised pre-training paradigm necessitates extensive labelled data for achieving effective pre-trained weights. 
Nonetheless, acquiring such datasets is costly and demands substantial professional expertise. 
Recently, the self-supervised pre-training paradigm has been receiving much attention from both the computer vision \cite{MoCo} and the remote sensing community \cite{SeCo}. 
This paradigm could acquire essential visual representations by constructing label-independent pretext tasks such as instance discrimination \cite{SeCo, liu2022multi, GeoAware}, spatial coherence \cite{Tile2Vec}, and masked image modeling \cite{he2022semantic}. 
Contrastive learning via instance discrimination is the most popular self-supervised method. 
The key to contrastive learning is to construct positive and negative example pairs. 
Manas et al. \cite{SeCo} explored utilizing the seasonal variation to construct the seasonal contrastive pairs. 
Ayush et al. \cite{GeoAware} utilized geographical location information and time changes to establish the contrastive learning objective. 
Liu et al. \cite{liu2022multi} introduced the consistency of SAR and optical image to realize contrast learning without negative samples. 

Pre-trained models are believed to be able to boost the downstream tasks; however, recent studies \cite{Wang_RS_Pretrain} have shown that general pre-training methodology may not be suitable for RSP. 
The suboptimal performance is observed when applying the pre-trained model to the downstream tasks, such as the segmentation task compared to IMP models \cite{Wang_RS_Pretrain}. 
The reason might be that IMP models can obtain diverse low-mid level features from natural images \cite{zhao2020makes}. 
These features play a pivotal role in dense prediction tasks, including semantic segmentation and object detection, possessing stronger transferability \cite{zhao2020makes} and serving as general knowledge.
Our experiments and previous research \cite{corley2023revisiting} consistently demonstrate that the IMP model can serve as a robust baseline for various remote sensing downstream tasks. 
Additionally, the differences in semantics between natural and remote sensing images can prevent semantic over-fitting \cite{ericsson2021well, wang2022RS_self}, further enhancing transferability. 

In contrast, RS images predominantly emphasize objects or scenes on the earth's surface, such as cars, houses, lakes, and airports. 
Additionally, they are constrained by the bird's-eye view perspective and sensor resolution, thereby restricting the diversity of scenes, perspectives, and detailed object information \cite{TOV}, thus impeding the learning of diverse low-mid level features. 
Simply scaling up the dataset does not bring more information enrichment. 
Furthermore, high-level semantic feature alignment in contrastive learning within RSP neglects the learning of low-mid level features \cite{wang2022revisiting}, thus diminishing performance on dense prediction tasks. 
IMP can efficiently acquire these features, prompting us to explore simultaneous pre-training using both RS and natural images.

In this study, we tackle this challenge by leveraging the rich knowledge of natural images to boost RSP. 
Introducing natural images to the RS domain enriches the feature space of RS pre-training models. 
To achieve this goal, training an IMP model on the RS images to acquire domain-specific knowledge \cite{CSPT} is one straightforward approach.
However, this multi-stage training procedure makes the model tend to forget the knowledge gained from the IMP phase, which hinders the pre-trained model from achieving satisfactory results on downstream tasks, as confirmed by our experiments.

To compensate for the shortcomings of the existing pre-training paradigms, we propose \textbf{Ge}neric Knowledge Boosted \textbf{R}emote \textbf{S}ensing \textbf{P}re-training (GeRSP) to obtain generic and remote sensing domain knowledge, as shown in Fig. \ref{fig:GeRSP_motivation}. 
In particular, a supervised pre-training branch on natural images is used to obtain general knowledge for downstream tasks. 
To capture RS domain knowledge, a self-supervised pre-training branch on RS images is co-operated with the supervised pre-training branch on natural images so that the proposed GeRSP simultaneously learns domain-related features from RS images.

In summary, our contributions include the following:
\begin{enumerate}
\item A novel remote sensing pre-training framework, GeRSP, is proposed to learn robust representations for RS understanding tasks. 
GeRSP uses a teacher-student architecture to simultaneously learn representations with general and domain knowledge. 
\item GeRSP contains supervised pre-training and self-supervised pre-training stages: (1) The self-supervised pre-training stage learns domain-related features from unlabeled RS images. (2) The supervised pre-training stage is integrated for general knowledge learning from labeled natural images. 
\item Three RS downstream tasks are evaluated to compare GeRSP with other pre-training methods, including object detection, semantic segmentation, and scene classification. The experimental results consistently demonstrate that GeRSP effectively improves the performance of remote sensing downstream tasks. 
Additionally, we perform a visual analysis to further evaluate the effectiveness of GeRSP and its impact on downstream tasks.
\end{enumerate}

\section{Related Work}
\subsection{Pre-training Methods}
Motivated by the observation that humans can leverage existing knowledge to solve new problems \cite{Pretrained}, transfer learning has been proposed as a solution to this challenge. 
Transfer learning allows models to benefit from pre-existing knowledge by leveraging pre-trained parameters. 
This idea of parameter transfer has been widely used in computer vision tasks.
There are two general pre-training methods based on whether labeled data is required during the training process: supervised pre-training and self-supervised pre-training. 

Supervised pre-training is a practical approach for obtaining pre-trained models. 
Models with different architectures, such as ResNet\cite{ResNet}, ViT \cite{ViT}, and Swin Transfromer \cite{Swin}, have been successfully pre-trained on large-scale image datasets like ImageNet \cite{ImageNet}. 
Benefiting from the general visual knowledge obtained from large-scale image datasets, downstream models only require a small amount of task-specific data to perform well. 
However, collecting and annotating large-scale datasets are still time-consuming and expensive in the real world, which limits the development of supervised pre-training methods. 
Semi-supervised learning \cite{chen2020big, berthelot2019mixmatch} can effectively leverage unlabeled data; however, it is often oriented towards specific task improvements and is less employed for pre-training.

Self-supervised pre-training methods are proposed to address these issues.
These methods construct pretext tasks that leverage intrinsic properties of images, thereby facilitating effective feature extraction \cite{SelfSupervised}. 
The constraint of pretext tasks compels the neural network to extract pertinent image information and generate good visual representations. 
The self-supervised pretext tasks encompass diverse methodologies, such as image generation \cite{GAN}, image inpainting \cite{ContextEncoder, MAE}, jigsaw puzzles \cite{Jigsaw}, and image colorization \cite{zhang2016colorful}. 
These tasks are designed based on the inherent structure and characteristics of the images themselves. 

Jing et al. \cite{SelfSupervised} summarized the pretext tasks into four categories: generation-based, context-based, free semantic label-based, and cross-modal-based. 
Generation-based methods encompass tasks such as image generation and image inpainting. 
Among them, methods based on image generation are primarily utilized for generating more realistic images \cite{BigGAN} or expanding datasets \cite{DAGAN} rather than focusing on acquiring a robust feature extractor. 
Pre-training methods based on image inpainting have gained significant traction within self-supervised pre-training methods, such as MAE \cite{MAE} and CAE \cite{CAE}. 
These methods involve masking specific regions within an image and requiring the network to predict the content of the masked areas. 
These approaches can effectively benefit downstream tasks that demand semantic understanding by necessitating the network's comprehension of the remaining image blocks and enabling image reconstruction based on contextual cues. 

The context-based pretext tasks primarily rely on semantic consistency or spatial context cues within the image as the supervisory signal. 
The pre-training method based on contrastive learning \cite{MoCo, MOCOv2, MOCOv3, SimCLR, BYOL} has gained significant traction among these tasks.  
The core idea behind contrastive learning is to minimize the distance between differently augmented views of the same image while maximizing the dissimilarity between unrelated images \cite{SimCLR, MoCo}. 
In our work, we incorporate contrastive learning into the remote sensing pre-training. 
Specifically, we employ MoCo technique \cite{MoCo} due to its widespread popularity, code base availability, and results reproducibility.

\subsection{Remote Sensing Pre-training Methods} 
Similar to other computer vision domains, ImageNet pre-trained models have demonstrated remarkable success in RS image recognition tasks \cite{hu2015transferring, marmanis2015deep, NWPU_RESISC45, RoITrans, ORCNN, HDMaps, jung2021boundary}. 
Nevertheless, challenges such as the domain gap between natural scenes and RS scenes and limitations in generalization persist in RS pre-training methods. 
Tong et al. \cite{tong2020land} presented a transfer learning approach for scene classification. 
Initially, the model is pre-trained with a well-annotated large-scale dataset. 
Subsequently, a semi-supervised transfer learning method is employed on the pre-trained model to achieve pixel-level classification. 
Building upon this methodology, Long et al. \cite{MAID} introduced Million-AID, a substantial benchmark dataset comprising one million instances designed explicitly for scene classification. 
The Million-AID dataset contains globally distributed high spatial resolution RS images in 51 scene categories. 
After that, a hierarchical multi-task learning framework \cite{MAID_Benchmark} for pixel-level scene classification demonstrates the strong generalization ability of Million-AID. 
In a complementary study, Wang et al. \cite{Wang_RS_Pretrain} conducted extensive experiments to assess the generalization performance of Million-AID pre-training models across multiple downstream tasks. 
The models employed encompass CNN-based architectures such as ResNet \cite{ResNet}, as well as Transformer-based approaches including Swin Transformer \cite{Swin} and ViTAEv2 \cite{Vitaev2}. 
The experimental results highlight that the pre-training models significantly enhance performance in various downstream tasks compared to the ImageNet pre-trained models. 
However, they also found that only using RS data may lack crucial information for detection and segmentation \cite{Wang_RS_Pretrain}, which inspired our investigation. 

Self-supervised pre-training methods have been extensively studied in RS research communities \cite{wang2022RS_self}. 
Numerous studies have employed pre-training methods to enhance the performance of specific RS tasks, such as hyperspectral imagery classification \cite{zhao2022hyperspectral, hou2021hyperspectral, zhu2021sc, RS_CMC}, synthetic aperture radar (SAR) target recognition \cite{wen2021rotation}, and change detection \cite{chen2021self, ou2022hyperspectral}. 
Some studies leveraged the geographic information associated with RS images to achieve more effective self-supervised pre-training. 
Jean et al. \cite{Tile2Vec} proposed Tile2Vec, an unsupervised representation learning method inspired by Word2Vec \cite{Word2Vec}. 
Tile2Vec assumes that geographically proximate tiles exhibit semantic similarity. 
Based on this assumption, they employed metric learning for unsupervised tiles learning. 

Jung et al. \cite{Remote_SmoothSimCLR} proposed a contrastive learning method based on the SimCLR \cite{SimCLR} framework. 
The method uses the idea of Tile2Vec \cite{Tile2Vec}, utilizing three neighbor tiles to obtain the smooth representation for positive samples. 
Ayush et al. \cite{GeoAware} exploited the revisiting characteristics of satellites to construct spatial-aligned image pairs at different times, enabling informative learning. 
Additionally, a geo-location pretext task was incorporated during training to enhance the representation learning of RS images. 
SeCo \cite{SeCo} combined temporal variation with other augmentation techniques to enable multi-augmentation contrastive learning. 
This approach yields representations encompassing time-varying and invariant features, offering advantages for downstream tasks. 
Scheibenreif et al. \cite{scheibenreif2022self} addressed land cover classification and segmentation tasks by employing SimCLR \cite{SimCLR} with paired satellite data obtained from optical Sentinel-2 and SAR Sentinel-1 sensors. 
However, it is essential to note that these studies require additional meta-information, including geographical location and time, which imposes more stringent restrictions on their practical application. 

While there has been considerable research on supervised and self-supervised pre-training models in the remote sensing domain, the comprehensive exploration of a general RS model with extensive generalization performance still needs to be improved. 
Risojevi{'c} et al. \cite{DAPT} proposed a domain-adaptive pre-training method that re-trains an ImageNet pre-trained model using MLRSNet \cite{MLRSNet} dataset. 
Their approach outperformed models pre-trained solely on either ImageNet \cite{ImageNet} or MLRSNet \cite{MLRSNet} in scene classification tasks.
Likewise, Zhang et al. \cite{CSPT} introduced the ConSecutive pre-training (CSPT) method for RSP. 
In CPST, the model first performs self-supervised learning on natural scene images through masked image modeling \cite{MAE} pretext task and then conducts self-supervised training on task-related unlabeled RS data. 
CSPT aims to bridge the domain gap and transfer knowledge from the natural image domain to the RS domain. 
However, it does not guarantee the preservation of general features learned from natural images during the second-stage self-supervised learning \cite{CSSL}. 
Additionally, the substantial size of ViT-based models restricts their practicality in certain situations. 

Our approach aims to improve the extraction of low-level general knowledge in RSP by incorporating supervised training with natural images, thereby enhancing spatial information perception capabilities. 
The motivation of TOV \cite{TOV} is close to ours.
They freeze natural image pre-trained model's shallow and middle layers and subsequently train on the RS dataset. 
This two-stage approach prevents the forgetting of general knowledge and achieves adaptability to remote sensing images. 
Compared with TOV, our method employs a joint training framework and facilitates the adaptation of even the shallow layer to remote sensing images. 
Furthermore, our approach demonstrates that simply introducing supervised learning with ImageNet \cite{ImageNet} can effectively acquire general knowledge without requiring a redundant multi-stage training strategy. 

\begin{figure*}[!htb]
	\centering
	\includegraphics[width=1\linewidth]{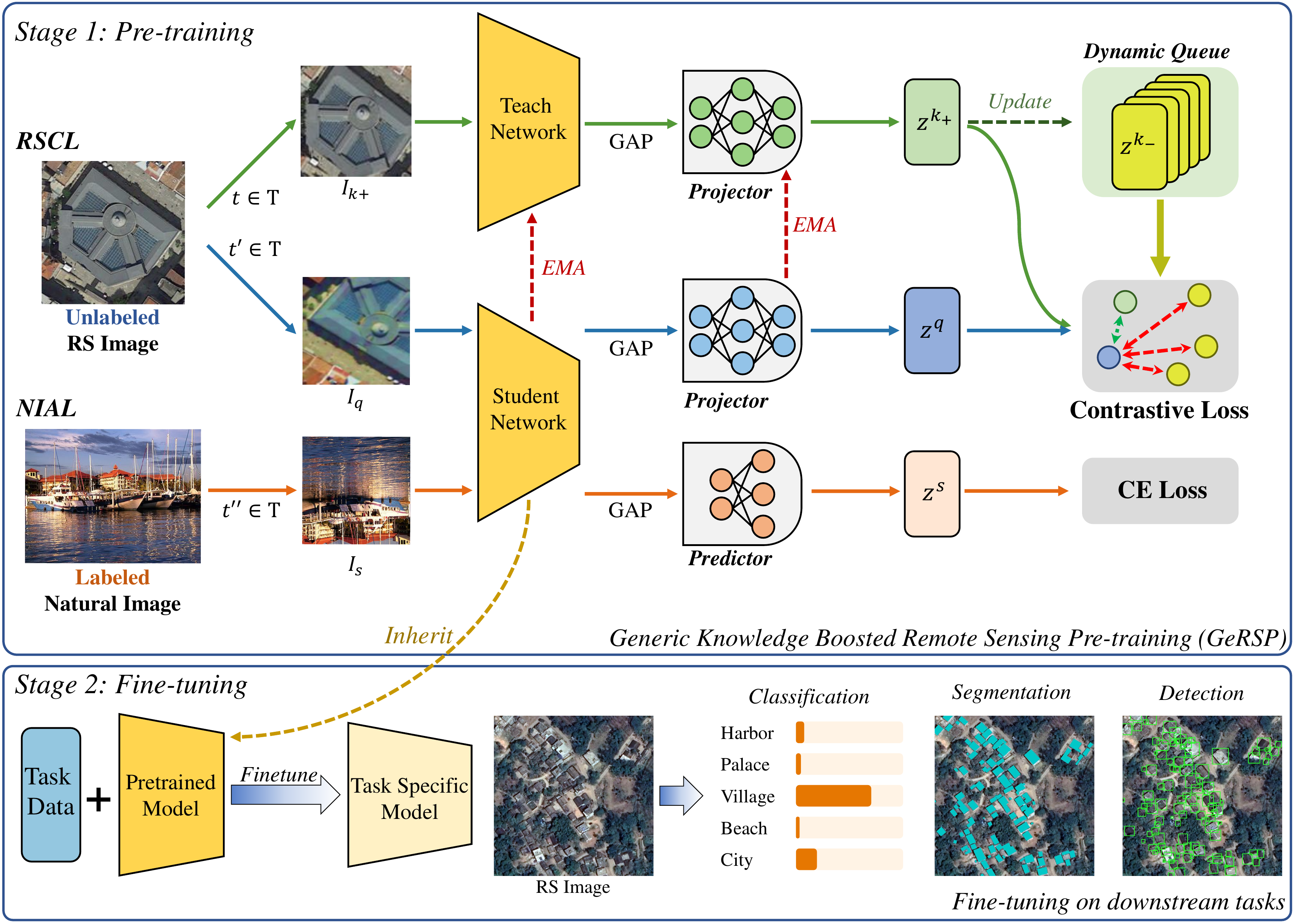}
	
	\caption{The overall framework of our proposed Generic Knowledge Boosted Remote Sensing Pre-training (GeRSP). GeRSP integrates two learning processes: natural image auxiliary learning (NIAL) on labeled natural images and remote sensing contrastive learning (RSCL) on unlabeled RS images. NIAL utilizes labeled natural images for training. NIAL involves training the model using labeled natural images, while RSCL adopts a contrastive learning approach. The trained model is subsequently fine-tuned on various downstream tasks using task-specific data.}
	\label{fig:method}
\end{figure*}
\section{Generic Knowledge Boosted Pre-training}

\begin{figure}[!htb]
     \centering
    \includegraphics[width=\linewidth]{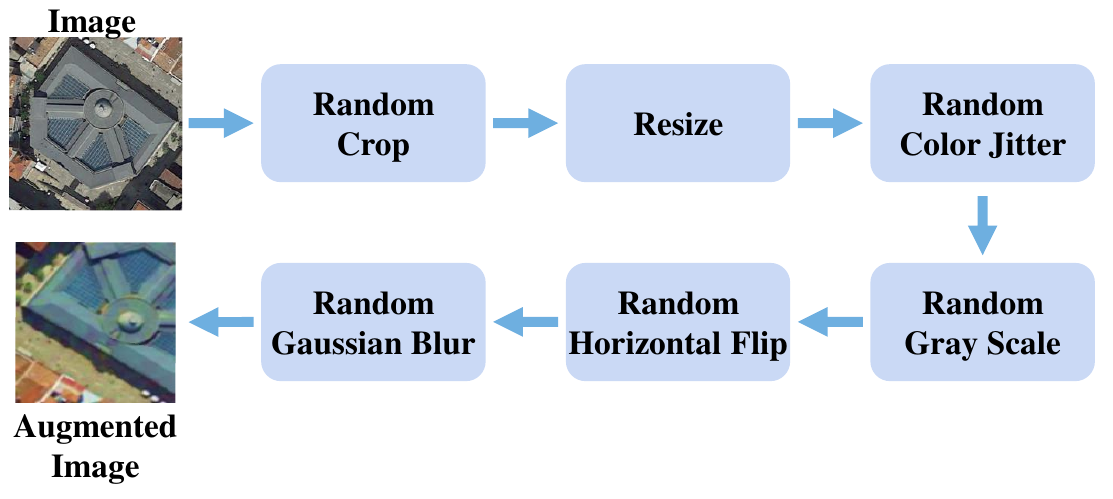}
	\caption{Data Augmentation Pipeline for pre-training.}
	\label{fig:SDA}
\end{figure}

\subsection{Overview}
Recently, remote sensing pre-training has established its effectiveness in extracting information from remote sensing data, with subsequent applicability to a broad spectrum of downstream tasks \cite{SeCo, GeoAware, RS_CMC, liu2022multi}. 
Simultaneously, ImageNet pre-training persists as a strong baseline due to its adeptness in assimilating transferrable knowledge from extensive large-scale natural data \cite{Wang_RS_Pretrain, corley2023revisiting}. 
Nevertheless, optimizing the concurrent utilization of both datasets for acquiring more robust transferable features remains a topic of ongoing research. 
To fill this gap, we propose GeRSP, a novel pre-training framework incorporating IMP and RSP. 
GeRSP learns the fundamental knowledge from IMP and captures specific knowledge tailored explicitly to remote sensing imagery. 
The downstream tasks employ GeRSP pre-trained weights as initialization weights for fine-tuning. This two-stage training methodology is illustrated in Fig. \ref{fig:method}.

As illustrated in Fig. \ref{fig:method}, GeRSP utilizes a teacher-student architecture to enable collaborative learning. 
It comprises two learning processes: natural image auxiliary learning (NIAL) on labeled natural images and remote sensing contrastive learning (RSCL) on unlabeled RS images. 
RSCL simultaneously trains both the teacher network and the student network. 
During the training process, the teacher network's parameters are updated using the exponential moving average of the student network parameters. 
Conversely, in the case of NIAL, the training is exclusively focused on the student network. 
Ultimately, the student network serves as a pre-trained model for downstream tasks. 
In each iteration of GeRSP, an equal number of natural and remote sensing images are sampled from their respective datasets. 

During RSCL, the RS image is subjected to two distinct augmentation strategies, denoted as $t$ and $t'$, to form positive pairs. 
Subsequently, the RS image pairs are fed into the teacher and student networks to extract their features. 
The features are then aggregated using global average pooling (GAP) and projected by independent \emph{projectors} that consists of two fully-connected layers, yielding the features $z^{k+}$ and $z^q$. 
Finally, the student network is optimized by minimizing the InfoNCE \cite{CPC} loss function computed over the positive sample pairs $(z^{k+}, z^q)$ and negative samples ${z^{k-}}$ for contrastive learning, where ${z^{k-}}$ are retrieved from a \emph{dynamic queue} that is actively maintained by the teacher network. 
Further details will be discussed in the subsequent subsection. 

Concurrently, labeled natural images from the same batch are utilized for NIAL to ensure adaptability to a wide range of tasks. 
Data augmentation is also applied to unlabeled natural images, resulting in the augmented images denoted as $I_s$. 
Then, the augmented images are inputted into the student network, followed by a GAP operation and an \emph{predictor}. 
The \emph{predictor}, distinct from the \emph{projector} used in RSCL, helps alleviate conflicts between NIAL and RSCL. 
The cross-entropy loss is then used to optimize the student network.
After each optimization iteration, both the parameters of the teacher network and the dynamic queue are updated. 
Subsequent sections will detail each component, including data augmentation (Sect. \ref{sec:DataAug}), the backbone network, projectors, and predictors for feature extraction in NIAL and RSCL (Sect. \ref{sec:Backbone}), the loss functions (Sect. \ref{sec:Loss}), and the update strategies for parameters and the dynamic queue (Sect. \ref{sec:update}).

\subsection{Data Augmentation}
\label{sec:DataAug} 

Data augmentation introduces variability in the input images, aiming at increasing the difficulty of the contrastive learning pretext task. 
Thus, the pre-trained model can acquire more meaningful features from augmented images rather than merely memorizing the input images \cite{MoCo, SimCLR, BYOL}. 
Therefore, the proposed GeRSP framework adopts the strong augmentation strategy described in \cite{StrongDataAug} for both RSCL and NIAL, enhancing the transferability of representations pre-trained on remote sensing (RS) images. 
Besides, the pre-trained model becomes more relevant to RSCL by employing strong data augmentation during NIAL, enabling more efficient feature learning. 

The pipeline for strong augmentation is depicted in Fig. \ref{fig:SDA}. 
Initially, images are cropped from the corresponding original image with a ratio ranging from 0.2 to 1 and then resized to a scale of 224. 
Subsequently, color jitter is applied with a probability of 0.8, employing brightness, contrast, saturation, and hue factors of 0.4, 0.4, 0.4, and 0.16, respectively. 
Moreover, images are converted to gray-scale with a probability of 0.2. 
Finally, images are flipped with a probability of 0.5 and subjected to Gaussian blur with a probability of 0.5. 

\subsection{Feature Extractor}
\label{sec:Backbone} 
Considering the practicality of the pre-trained model \cite{SeCo}, we select ResNet50 \cite{ResNet} as the backbone network for pre-training.
During RSCL, we employ shuffling batch normalization (BN) \cite{MoCo} in the backbone to eliminate the correlation between BN parameters and the mini-batch, which can effectively avoid information leakage. 
Specifically, the order of samples within the mini-batch is shuffled before input to the backbone network.
The original order is recorded to be restored during contrastive learning. 

After the backbone network has extracted image features, Global Average Pooling (GAP) is applied to reduce the dimensionality and obtain features in $\mathbb{R}^{2048}$.
As illustrated in Fig. \ref{fig:method}, the non-linear \emph{projector} and \emph{predictor} are employed to process these features further. 

The \emph{projector} consists of two fully connected layers with ReLU activation. 
It has a hidden dimension of $2048$ and an output dimension of $128$, yielding the features $z^{k+}$ and $z^q$. 
On the other hand, the \emph{predictor} is a single fully connected layer that maps features to logits $z^{s}$ for classification. 
The introduction of non-linearity in the \emph{projector} can prevent dimension collapse \cite{DirectCLR} and enhance the performance of multitask learning. 
Since RSCL aims to achieve feature invariance by strong augmentation in input images, it can adversely affect task-specific characteristics learning such as image color, contrast, and position. 
Therefore, the \emph{projector} is crucial for RSCL and can learn semantic invariance features while preserving more information for the backbone network. 
Additionally, the \emph{projector} serves as a bridge to close the information gap between NIAL and RSCL. 

\subsection{Loss Function}
\label{sec:Loss}
The loss function for GeRSP consists of two terms: the cross-entropy loss for NIAL, denoted as $L_{CE}$, and the contrastive loss for RSCL, denoted as $L_{CT}$. 
For $L_{CE}$, we firstly normalize $z^s$ by using the softmax function, which yields a probability vector $\mathbf{p}=(p_1, p_2, ..., p_K)$, where $K$ represents the number of categories. 
Then, the cross-entropy loss $L_{CE}$ is computed by using the following equation: 
\begin{equation}
L_{CE}(\mathbf{p}, \mathbf{y}) = - \sum_{i=1}^{K} y_i \log(p_i) 
\label{eq:CE_loss}
\end{equation}
where $y_i$ represents the ground truth label for the corresponding category. 
For $L_{CT}$, RSCL utilizes the InfoNCE loss as the contrastive loss function. 
Considering features $z^{k_+}$ and $z^q$ obtained from the \emph{projectors}, as well as features derived from the negative queue comprising historical features $\mathbf{Z}^{k_-}={z^{k_-}}$, the calculation of the contrastive loss \cite{CPC} is as follows: 
\begin{equation}
L_{CT}(z^q, \mathbf{Z}^{k_-}) = -\log\frac{\exp(s_{q,k_+}/\tau)}
{\exp(s_{q,k+}/\tau) + 
\sum\limits_{k_-}{\exp(s_{q,k_-}/\tau)}}
\label{eq:CT_Loss}
\end{equation}
where $\tau$ is the temperature parameter that controls the intensity of contrast, $s_{q, k} = z^q \cdot z^k / || z^q|| || z^k||$ is the cosine similarity between features. 
The cross-entropy loss used in NIAL is actually similar to the InfoNCE loss \cite{MoCo} employed in RSCL, allowing for the joint training of NIAL and RSCL. 
The total loss is defined as below: 
\begin{equation}
L_{total} = L_{ct} + \alpha L_{ce}
\label{eq:TotalLoss}
\end{equation}
where coefficient $\alpha$ is employed to strike a balance between NIAL and RSCL, with a default value of 1. 

\subsection{Parameter \& Queue Update}
\label{sec:update}
The teacher network employs momentum update \cite{MoCo} on the network parameters to facilitate a stable training process.
Denoting parameters of the teacher network and the student network as $W_{t}$ and $W_{s}$ respectively, the update rule for $W_t$ is as follows: 
\begin{equation}
W_t = mW_t + (1-m)W_s
\label{eq:MomentumUpdate}
\end{equation}
where the momentum coefficient $m$ is set to 0.996. 
The dynamic queue is implemented as a First-In-First-Out (FIFO) queue with a maximum capacity of 65,536. 
It is updated after each iteration and serves as a storage for features generated by the teacher network. 
Given a batch size of 128, it takes approximately 500 iterations to complete a full update of the queue.

\section{pre-training}
\subsection{Implementation Details}
To compare the effectiveness of our proposed GeRSP with other pre-training methods, we use ResNet50 \cite{ResNet} as the backbone and select several pre-training methods, such as supervised pre-training and self-supervised pre-training. 
To validate the superiority of GeRSP over RSP, we choose SeCo \cite{SeCo}, GeoAware \cite{GeoAware}, CACO \cite{CACO}, TOV \cite{TOV}, MoCo \cite{MoCo}, and MoCov2 \cite{MOCOv2} as comparative methods. 
To demonstrate the advantages of GeRSP over IMP, we compare it with supervised pre-training and MoCo \cite{MoCo}. 
Furthermore, we conduct training using MoCo on mixed data as a stronger baseline to show that GeRSP can better utilize mixed data.

The unlabeled RS image dataset Million-AID (MAID) \cite{MAID} and the labeled natural image dataset ImageNet \cite{ImageNet} are utilized for pre-training. 
MAID\cite{MAID} is a large-scale remote sensing scene classification dataset consisting of 1,000,848 images with 51 scene categories.
MAID is collected from Google Earth with broad resolutions, ranging from 0.5 to 153 m per pixel. 

All pre-training methods are trained with the stochastic gradient descent (SGD) optimizer for 100 epochs, with initial values of 0.05 for the learning rate, 0.90 for weight decay, and 0.00005 for momentum.
The learning rate for GeRSP is optimized using the cosine annealing scheduler with restarts \cite{SGDR}: 
\begin{equation}
lr_{cur} = lr_{min} + \frac{1}{2}(lr_{max} - lr_{min})(1+\cos(\frac{T_{cur}}{T_{max}}\pi))
\label{cosine}
\end{equation}
The minimum learning rate, denoted as $lr_{min}$, is set to 0.10, while the maximum learning rate, denoted as $lr_{max}$, is set to 0.01. 
During a single round of training, $T_{max}$ is defined as the maximum number of epochs, which is set to 20. 
$T_{cur}$ represents the current epoch number within the current round. 
Once $T_{cur}$ reaches $T_{max}$, it is reset to 0 to initiate the subsequent round, and $lr_{cur}$ denotes the current learning rate. 

For MoCo and MoCov2, learning rates are optimized using the step linear scheduler with a step size of 30 and a learning rate decay of 0.1. 
Pre-training methods are trained in a distributed manner, utilizing data parallelism across 8 RTX-2080 GPUs, with a total batch size of 128. 
After the pre-training stage, irrelevant components, including the predictor, projector, and the teacher network, are removed, and the student network acts as the pre-trained model for downstream tasks. 
For previous RSP methods (SeCo, GeoAware, CACO, and TOV), we directly download their pre-trained parameters and evaluate them on downstream tasks to objectively compare the performance of each method.

\section{Finetuning On Downstream Tasks}
To evaluate the effectiveness of pre-training methods, we finetune different pre-trained models on three downstream tasks: scene classification, object detection, and semantic segmentation. 
All pre-trained models in these tasks adhere to the same configurations and hyper-parameter settings.

\subsection{Scene Classification}
To align pre-trained models with the requirements of classification tasks, they undergo augmentation by adding an average pooling layer and a single fully connected layer at the final stage. 
The models are optimized by mini-batch stochastic gradient descent with momentum (SGDM) algorithm for 100 epochs, with a batch size of 64. 
For SGDM, the initial learning rate, weight decay, and momentum are set to 0.01, 0.0001, and 0.9, respectively.
A linear-step scheduler is utilized to ensure training stability, employing step values of [30, 60, 90] and a decay ratio of 0.1. 
For image pre-processing, the images are scaled to $224\times224$ pixels. Additionally, random flipping is applied with a probability of 0.5, meaning each image has a 50\% 
Two scene classification datasets are employed for validation:
\begin{enumerate}
\item[$\bullet$] EuroSAT \cite{EuroSAT}: This dataset consists of Sentinel-2 satellite images captured over Europe. 
It consists of 27,000 images belonging to 10 distinct categories. 
Each class comprises approximately 2,000 to 3,000 images, with a resolution of $64\times64$ pixels. 

\item[$\bullet$] NWPU-RESISC45 \cite{NWPU_RESISC45}: This dataset, developed by Northwestern Polytechnical University (NWPU) for Remote Sensing Image Scene Classification (RESISC), includes 31,500 images. 
The dataset covers 45 scene categories, with 700 images per category. 
The images have spatial resolutions ranging from 30 to 0.2 meters, and their size is $256\times256$ pixels. 
\end{enumerate}

The evaluation metric employed for classification accuracy is the Top-1 accuracy. 
Following \cite{Wang_RS_Pretrain}, we employ 20\% of the data for training and reserved 80\% for testing. 
The models are repeatedly trained and evaluated five times at each setting. The average value $\mu$ and standard deviation $\sigma$ of the results across various trials were documented as $\mu$ ± $\sigma$.

\subsection{Object Detection}
To assess the performance of pre-trained models, we employ three detection methods, e.g., Faster R-CNN \cite{Faster-RCNN}, RetinaNet \cite{RetinaNet}, and Dy-Head \cite{Dy-Head}. 
These detection methods are implemented using MMDetection \cite{mmdetection} toolbox. 
The pre-trained backbone parameters are replaced with the parameters obtained from our experiments. 
The detectors are trained using the SGDM optimizer with a learning rate of 0.001, momentum of 0.90, and weight decay of 0.0001. 
The learning rate is reduced by a factor of 10 at 16 epochs and 22 epochs, respectively. 
The experiments are conducted on two GPUs with a batch size of 4 over 24 epochs. 
During training and testing, images are resized to 800$\times$800 pixels. 
Only random flipping is applied during training. 
For testing, non-maximum suppression (NMS) with an intersection over union (IoU) threshold of 0.3 is employed to remove duplicated detections and retains a maximum of 1,000 detections. 
The widely used DOTA \cite{DOTA} and DIOR \cite{DIOR} datasets are selected in our experiments: 
\begin{enumerate}
\item[$\bullet$] DOTA \cite{DOTA}: This dataset comprises 2,806 images with a total of 188,282 instances belonging to 15 different categories. 
The images have varying sizes ranging from $800\times800$ pixels to 4,000$\times$4,000 pixels. 
In our experiment, we utilize horizontal bounding box (HBB) annotations, which require minimal modifications to enable Faster R-CNN for RS object detection. 
The images are cropped into patches of size $800\times800$ pixels with a stride of 640. 
The performance on the cropped validation set is reported. 
\item[$\bullet$] DIOR \cite{DIOR}: The DIOR dataset consists of 23,463 images with a total of 192,472 instances. 
It covers 20 object categories and offers a significantly more diverse distribution of instances and finer classification than other datasets. 
\end{enumerate}

\begin{table}[t!]
  \centering
  \caption{The comparison of different pre-training methods on scene classification task.}
  \setlength\tabcolsep{2pt} 
\begin{tabular}{c|c|c|c}
    \toprule
    \multirow{2}[2]{*}{Pre-training} & \multirow{2}[2]{*}{Dataset} & EuroSAT & NWPU-RESISC45 \\
          &       & Top-1 & Top-1 \\
    \midrule
    IMP   & ImageNet & 97.84 ± 0.04 & 92.48 ± 0.13 \\
    MoCo-IN \cite{MoCo} & ImageNet & 97.55 ± 0.29 & 90.77 ± 0.19 \\
    MoCo-MAID \cite{MoCo} & MAID  & 97.52 ± 0.10 & 91.19 ± 0.15 \\
    MoCo-IN-MAID \cite{MoCo} & ImageNet + MAID & 97.47 ± 0.10 & 91.15 ± 0.20 \\
    MoCov2 \cite{MOCOv2} & MAID  & 97.52 ± 0.07 & 91.28 ± 0.04 \\
    SeCo \cite{SeCo} & 1M Sentinel-2 & 97.74 ± 0.25 & 90.40 ± 0.15 \\
    Geo-Aware \cite{GeoAware} & GeoImageNet & 97.87 ± 0.11 & 92.37 ± 0.19 \\
    CACO \cite{CACO} & CACO 1M &  97.67 ± 0.09  &  90.81 ± 0.18  \\
    TOV \cite{TOV} & TOV-NI + TOV-RS &  97.80 ± 0.06  &  92.59 ± 0.21  \\
    \textbf{GeRSP} & ImageNet + MAID & \textbf{97.87 ± 0.15} & 92.67 ± 0.16 \\
    \textbf{GeRSP-200} & ImageNet + MAID & \textbf{97.87 ± 0.10} & \textbf{92.74 ± 0.09} \\
    \bottomrule
\end{tabular}%
  \label{tab:classification}%
\end{table}%

\begin{table*}[htbp]
  \centering
 \setlength\tabcolsep{3pt}
 \renewcommand{\arraystretch}{1.05} 

  \caption{Fine-tuning results on the DIOR \cite{DIOR} object detection task. 
  We use three detection methods to verify the pre-trained models.}
    \begin{tabular}{ccccccccccccccccccccc|c}
    \toprule
    Methods & APL   & APO   & BF    & BC    & BR    & CH    & DAM   & ETS   & ESA   & GF    & GTF   & HA    & OP    & SH    & STA   & STO   & TC    & TS    & VE    & WM    & mAP \\
    \midrule
    \textit{Faster R-CNN} &       &       &       &       &       &       &       &       &       &       &       &       &       &       &       &       &       &       &       &       &  \\
    From Scratch & 45.4  & 28.2  & 63.1  & 59.9  & 18.7  & 57.4  & 32.1  & 38.1  & 43.4  & 40.2  & 48.6  & 40.4  & 37.0  & 69.8  & 45.7  & 43.5  & 75.1  & 27.9  & 31.8  & 70.0  & 45.8  \\
    IMP   & 58.6  & 74.0  & 70.9  & 86.0  & 38.9  & 77.5  & 55.2  & 59.9  & 69.4  & 74.2  & \textbf{80.4} & 51.3  & 55.5  & 75.1  & 62.9  & 57.0  & 83.6  & 50.2  & 39.6  & 81.9  & 65.1  \\
    MoCo-IN \cite{MoCo} & 53.8  & 73.5  & 67.6  & 86.1  & 41.8  & 75.4  & 55.3  & 61.9  & 73.2  & 74.9  & 77.2  & \textbf{57.6} & 57.3  & \textbf{77.2} & 67.7  & 55.4  & 84.6  & 56.2  & 39.2  & 82.3  & 65.9  \\
    MoCo-MAID \cite{MoCo} & 53.8  & 73.3  & 69.4  & 85.9  & 40.6  & 75.1  & 56.3  & 65.7  & 73.1  & 76.1  & 77.7  & 57.4  & 57.3  & 77.1  & 64.9  & 57.4  & 84.5  & 53.4  & 40.3  & \textbf{83.1} & 66.1  \\
    MoCo-IN-MAID \cite{MoCo} & 53.7  & 73.4  & 69.2  & 84.9  & 41.7  & 73.5  & 55.3  & 62.7  & 72.7  & 74.6  & 78.4  & 57.4  & 56.7  & 77.0  & 64.3  & 54.3  & 85.5  & \textbf{59.2} & 39.7  & 82.2  & 65.8  \\
    MoCov2 \cite{MOCOv2} & 50.9  & 72.7  & 69.2  & 85.8  & 41.6  & 74.0  & 57.1  & 62.4  & 70.6  & 75.0  & 78.2  & 56.8  & 56.5  & 77.1  & 65.2  & 54.3  & 84.6  & 54.8  & 39.7  & 83.0  & 65.5  \\
    SeCo \cite{SeCo} & 56.3  & 56.6  & 71.3  & 81.4  & 33.4  & 71.1  & 48.2  & 55.1  & 63.5  & 65.6  & 71.7  & 47.9  & 49.9  & 74.5  & 49.0  & 51.8  & 82.4  & 46.5  & 37.6  & 80.3  & 59.7  \\
    Geo-Aware \cite{GeoAware} & 52.7  & 68.4  & 73.3  & 83.7  & 36.6  & 75.5  & 50.4  & 63.0  & 68.1  & 69.5  & 75.9  & 47.7  & 53.5  & 74.8  & 60.2  & 57.6 & 85.0  & 46.8  & 39.6  & 80.7  & 63.1  \\
    CACO \cite{CACO} & 53.8  & 64.6  & 73.4 & 82.6  & 35.2  & 72.7  & 49.0  & 56.7  & 65.6  & 69.9  & 75.1  & 48.7  & 50.7  & 75.7  & 54.8  & 54.6  & 84.5  & 44.6  & 38.4  & 80.4  & 61.6  \\
    TOV \cite{TOV}  &  \textbf{63.9}  & 72.4  & \textbf{77.1} & 85.0  & 40.3  & 77.3  & 57.3  & 66.1  & 71.9  & 75.1  & 80.0  & 51.1  & 55.4  & 76.9  & 63.9  & \textbf{59.8} & 85.9  & 49.0  & 41.1  & 82.8  & 66.7  \\
    GeRSP & 61.5  & 76.2  & 69.1  & 86.7  & 42.3  & 77.6  & 59.9  & 62.4  & 72.8  & \textbf{77.6} & 78.4  & 54.5  & 58.4  & 75.3  & \textbf{68.4} & 57.2  & 86.3  & 55.1  & 41.0  & 82.3  & 67.1  \\
    GeRSP-200 & 63.6 & \textbf{78.1} & 69.4  & \textbf{87.1} & \textbf{44.0} & \textbf{78.5} & \textbf{62.9} & \textbf{67.0} & \textbf{76.0} & 76.8  & 79.6  & 54.7  & \textbf{58.7} & 75.3  & 60.3  & 57.3  & \textbf{86.4} & 56.3  & \textbf{41.3} & 80.4  & \textbf{67.8} \\
    \midrule
    \textit{RetinaNet} &       &       &       &       &       &       &       &       &       &       &       &       &       &       &       &       &       &       &       &       &  \\
    From Scratch & 51.1  & 37.9  & 64.4  & 59.0  & 19.8  & 59.4  & 38.0  & 43.5  & 54.3  & 51.0  & 58.2  & 45.3  & 39.7  & 68.9  & 61.9  & 37.5  & 77.6  & 27.8  & 27.8  & 67.0  & 49.5  \\
    IMP   & 66.6  & 77.4  & 74.5  & 87.2  & 35.8  & 79.9  & 59.4  & 55.6  & 75.4  & 80.4  & \textbf{79.7} & 49.5  & 54.8  & 75.8  & 68.8  & 52.7  & 85.5  & 50.1  & 39.6  & 82.4  & 66.5  \\
    MoCo-IN \cite{MoCo} & 57.1  & 70.8  & 71.0  & 85.3  & 35.8  & 74.8  & 54.2  & 51.4  & 70.4  & 74.8  & 74.9  & 49.9  & 54.4  & 77.1  & 59.6  & 54.5  & 84.8  & 46.5  & 39.7  & 82.5  & 63.5  \\
    MoCo-MAID \cite{MoCo} & 57.7  & 72.7  & 71.4  & 86.0  & 35.0  & 75.8  & 52.6  & 55.3  & 73.6  & 75.1  & 75.6  & 51.4  & 53.8  & 76.5  & 64.6  & 53.4  & 85.6  & 43.4  & 40.8  & 82.8  & 64.2  \\
    MoCo-IN-MAID \cite{MoCo}& 51.2  & 70.9  & 72.1  & 86.7  & 35.0  & 75.5  & 57.6  & 52.6  & 70.4  & 75.8  & 74.5  & 49.9  & 52.8  & 75.7  & 63.2  & 51.2  & 85.0  & 43.8  & 38.5  & 80.2  & 63.1  \\
    MoCov2 \cite{MOCOv2}& 54.1  & 71.8  & 70.9  & 86.7  & 35.3  & 74.7  & 59.3  & 52.1  & 70.2  & 74.8  & 74.7  & 51.0  & 53.8  & 75.7  & 61.5  & 51.9  & 85.0  & 42.1  & 39.9  & 82.3  & 63.4  \\
    SeCo \cite{SeCo} & 54.7  & 60.9  & 71.1  & 81.1  & 30.3  & 72.7  & 56.3  & 53.2  & 69.6  & 73.7  & 73.2  & 50.7  & 48.5  & 75.0  & 65.0  & 47.9  & 84.0  & 45.4  & 34.9  & 78.6  & 61.3  \\
    Geo-Aware \cite{GeoAware} & 60.4  & 67.9  & 71.8  & 83.6  & 31.3  & 77.4  & 51.9  & 55.1  & 71.5  & 72.2  & 75.1  & 46.6  & 50.9  & 73.9  & 62.3  & 51.6  & 85.7  & 42.4  & 37.6  & 78.1  & 62.4  \\
    CACO \cite{CACO} & 56.0  & 67.9  & 72.3  & 81.6  & 30.8  & 74.3  & 56.5  & 53.9  & 71.6  & 74.9  & 75.2  & 49.8  & 50.9  & 75.1  & 69.6  & 49.1  & 84.8  & 41.4  & 36.4  & 80.1  & 62.6  \\
    TOV \cite{TOV}   & 63.4  & 74.3  & 75.1  & 83.1  & 37.5  & 77.2  & 60.9  & 59.8  & 73.9  & \textbf{81.6} & 77.7  & 51.8  & 54.3  & 76.4  & 68.6  & 56.1  & 86.1  & 47.5  & 40.1  & 82.2  & 66.3  \\
    GeRSP & \textbf{70.3} & 76.9  & \textbf{75.3} & 87.2  & 39.4  & \textbf{80.3} & 61.4  & 59.3  & \textbf{77.3} & 80.5 & 76.8  & 50.5  & 56.0  & 76.8  & 64.3  & 54.3  & 86.8  & \textbf{50.3} & 42.2  & 83.8  & 67.5  \\
    GeRSP-200 & 70.0  & \textbf{77.7} & 74.5  & \textbf{87.8} & \textbf{41.2} & \textbf{80.3} & \textbf{63.6} & \textbf{61.9} & 76.9  & 80.3  & 78.7  & \textbf{51.9} & \textbf{56.8} & \textbf{78.1} & \textbf{72.6} & \textbf{57.0} & \textbf{87.7} & 50.0  & \textbf{43.0} & \textbf{84.6} & \textbf{68.7} \\
    \midrule
    \textit{DyHead} &       &       &       &       &       &       &       &       &       &       &       &       &       &       &       &       &       &       &       &       &  \\
    From Scratch & 58.6  & 61.1  & 70.5  & 68.6  & 28.3  & 65.3  & 46.4  & 50.5  & 65.2  & 63.2  & 62.9  & 51.8  & 46.5  & 77.6  & 66.5  & 47.5  & 78.9  & 50.1  & 35.4  & 75.7  & 58.5  \\
    IMP   & 63.8  & 82.6  & 76.4  & 86.0  & 43.2  & \textbf{79.5} & 65.0 & 67.5  & 78.7  & \textbf{80.2} & \textbf{79.0} & 57.0  & 58.6  & 82.7  & 67.8  & 64.9  & 86.3 & 60.6  & 46.6  & \textbf{86.6} & 70.6  \\
    MoCo-IN \cite{MoCo}& 64.1  & 78.7  & 74.9  & 85.8  & 43.0  & 74.7  & 59.4  & 65.8  & 75.4  & 77.1  & 77.4  & \textbf{59.4} & 57.9  & \textbf{85.9} & 70.9 & 67.5  & 85.6  & 59.1  & 46.9  & 85.5  & 69.8  \\
    MoCo-MAID \cite{MoCo} & 60.6  & 80.7  & 72.0  & 86.5  & 42.2  & 74.3  & 60.7  & 61.8  & 75.5  & 74.8  & 76.4  & 57.7  & 57.4  & 84.4  & 69.8  & 61.6  & 86.3  & 55.5  & 46.7  & 85.7  & 68.5  \\
    MoCo-IN-MAID \cite{MoCo} & 61.3  & 78.9  & 75.3  & 86.2  & 42.7  & 75.8  & 62.0  & 61.0  & 75.7  & 77.1  & 75.0  & 57.8  & 57.8  & 84.2  & 67.3  & 62.2  & 85.1  & 58.7  & 45.6  & 85.2  & 68.7  \\
    MoCov2 \cite{MOCOv2}& 64.0  & 80.1  & 74.8  & 86.4  & 44.0  & 74.9  & 61.1  & 61.7  & 75.8  & 75.1  & 76.3  & 59.3  & 59.4  & 85.0  & 70.2  & 64.4  & 85.0  & 55.5  & 47.3  & 86.4  & 69.3  \\
    SeCo \cite{SeCo}  & 57.5  & 75.2  & 73.9  & 81.3  & 37.4  & 72.9  & 61.6  & 60.0  & 72.5  & 74.2  & 72.5  & 56.6  & 53.6  & 80.4  & 66.3  & 57.1  & 82.9  & 58.6  & 41.4  & 83.0  & 66.0  \\
    Geo-Aware \cite{GeoAware}& 63.0  & 77.4  & 76.1  & 85.1  & 42.2  & 76.5  & 61.2  & 61.6  & 75.9  & 72.2  & 74.7  & 56.7  & 57.6  & 83.7  & 67.2  & 63.5  & 85.5  & 60.0  & 45.1  & 84.0  & 68.4  \\
    CACO \cite{CACO} & 59.4  & 76.5  & 75.3  & 83.2  & 40.9  & 73.3  & 58.3  & 62.6  & 73.8  & 75.7  & 73.5  & 56.5  & 55.9  & 83.0  & 69.7  & 58.7  & 83.6  & 54.6  & 43.0  & 82.9  & 67.0  \\
    TOV \cite{TOV} & 64.7  & 81.7  & \textbf{77.5} & 85.6  & 44.8  & 76.0  & \textbf{65.7}  & 64.9  & 79.6  & 77.8  & 77.6  & 57.1  & 57.2  & 81.7  & \textbf{71.1} & 65.3  & 84.8  & \textbf{63.3} & 44.8  & 85.0  & 70.2  \\
    GeRSP & 68.4  & 83.6  & 75.9  & \textbf{87.4} & 47.0  & 78.4  & 63.3  & 67.8  & 79.4  & 79.1  & 77.6  & 57.7  & 59.1  & 84.2  & 68.4  & 66.8  & 86.6  & 59.0  & 48.1  & 86.4  & 71.2  \\
    GeRSP-200 & \textbf{72.0} & \textbf{84.1} & 76.8 & 86.9  & \textbf{47.3} & 79.0  & 62.3  & \textbf{70.9} & \textbf{81.4} & 78.9  & 78.9  & 58.2  & \textbf{60.7} & 83.9  & 70.4  & \textbf{67.7} & \textbf{86.9} & 62.8 & \textbf{48.9} & \textbf{86.6} & \textbf{72.2} \\
    \bottomrule
    \end{tabular}%
  \label{tab:DIOR}%
\end{table*}%

During the evaluation, we compare the Average Precision (AP) of each category, and the mean Average Precision (mAP) is also considered. 
Specifically, we adopt the evaluation protocol of COCO \cite{COCO} and use the AP calculated under the IoU threshold of 0.5 as the evaluation criterion. 

\subsection{Segmentation}
Two classic segmentation methods (i.e., PSANet \cite{PSANet} and DeepLabV3+ \cite{DeepLabV3+}) are chosen in our experiments. 
The two segmentation models are all trained in 80,000 iterations with a batch size of 4, using the SGD algorithm with a learning rate of 0.01, weight decay of 0.0005, and momentum term of 0.9. 
During training, images are resized to 2,048$\times$512 pixels, randomly cropped to 512$\times$512 patches, and randomly flipped along the horizontal axis. 
For more accurate segmentation results, images are resized to 1,024$\times$1,024 during testing. 
High spatial resolution land-cover semantic segmentation is selected as the segmentation task on land-cover dataset LoveDA. 
\begin{enumerate}
\item[$\bullet$] LoveDA \cite{LoveDA}: The dataset contains 5,987 satellite images with 166,768 annotated objects from three cities: Nanjing, Changzhou, and Wuhan. LoveDA covers 536.15 $km^2$ and each image includes multi-scale objects, complex background, and inconsistent class distributions.
There are 2,522 and 1,669 images in the training and validation sets, respectively.
The typical resolution of images in the dataset is 1,024$\times$1,024.
\end{enumerate}

The mean IoU (mIoU) is chosen as the metric for evaluation. 
\section{EXPERIMENTS AND ANALYSIS} 
\subsection{Results} 
We selected ImageNet supervised pre-training (IMP), MoCov1 \cite{MoCo}, MoCov2 \cite{MOCOv2}, SeCo \cite{SeCo}, GeoAware \cite{GeoAware}, CACO \cite{CACO}, and TOV \cite{TOV} as comparison pre-training methods. ImageNet supervised pre-training and MoCov1 serve as the baseline for GeRSP. 
To illustrate the impact of the dataset on the pre-training method in downstream tasks, we pre-train the backbone network with MoCo on ImageNet, MAID, and ImageNet+MAID datasets, respectively. 
The obtained pre-trined models are denoted as MoCo-IN, MoCo-MAID, and MoCo-IN-MAID, respectively. 
To demonstrate the sustained effectiveness of GeRSP in larger-scale training, we conduct training for 200 epochs, referred to as GeRSP-200.
The results of semantic segmentation are shown in Table \ref{tab:classification}, Table \ref{tab:DIOR}, Table \ref{tab:DOTA} and Table \ref{tab:segmentation}, respectively. 

\textbf{Overall Performance:} 
Across all the obtained results, GeRSP consistently exhibited improvements. 
Notably, GeRSP delivers substantial enhancements in detection, attaining superior performance across all methods and datasets.
Moreover, GeRSP demonstrates the capability to enhance both segmentation and classification endeavors.
Compared to training from scratch, all pre-training methods prove efficacious in augmenting the performance of fine-tuning on downstream tasks, thereby emphasizing the indispensability of pre-training in remote sensing cognitive tasks.

\textbf{Scene Classification:}
Table \ref{tab:classification} reports the top-1 accuracy of our proposed GeRSP and other pre-training methods on EuroSAT and NWPU-RESISC45. 
As shown in Table \ref{tab:classification}, GeRSP outperforms the other pre-training methods on the two datasets, showing its effectiveness. 
For MoCo-IN, MoCo-MAID, and MoCo-IN-MAID, GeRSP improves the top-1 accuracy by +2.0\%, +1.6\%, and +1.6\% on NWPU-RESISC45, respectively. 
The experiments show domain gaps exist between RS images and natural images when comparing MoCo-IN with GeRSP. 
Moreover, pre-training only on RS images can lead to the lack of general features of objects, which reduces the generalization performance on downstream tasks when comparing MoCo-MAID with GeRSP. 
The proposed GeRSP, on the other hand, learns representations with general and special knowledge simultaneously through a unified framework. 
Compared with SeCo, CACO, TOV, and Geo-Aware, GeRSP consistently improves top-1 accuracy, indicating that GeRSP can learn robust representations with both general knowledge and domain specializations.

\begin{table*}[htbp]
 \setlength\tabcolsep{6.5pt}
\renewcommand{\arraystretch}{1.05} 
  \centering
  \caption{Fine-tuning results on the DOTA \cite{DOTA} object detection task. 
  We use three detection methods to verify the pre-trained models.}
    \begin{tabular}{cccccccccccccccc|c}
    Methods & PL    & BD    & BR    & GTF   & SV    & LV    & SH    & TC    & BC    & ST    & SBF   & RA    & HA    & SP    & HC    & mAP \\
    \midrule
    \textit{Faster R-CNN} &       &       &       &       &       &       &       &       &       &       &       &       &       &       &       &  \\
    From Scratch & 76.5  & 44.0  & 23.2  & 21.5  & 63.7  & 73.3  & 83.4  & 85.5  & 13.7  & 48.0  & 21.6  & 25.1  & 66.6  & 47.3  & 19.8  & 47.6  \\
    IMP   & 84.1  & 63.3  & 41.4  & \textbf{60.9} & 66.9  & \textbf{77.1} & 86.2  & 92.6 & 53.0  & 61.4  & 61.0  & 56.4  & 76.1  & \textbf{52.3} & 44.1  & 65.1  \\
    MoCo-IN \cite{MoCo} & 84.1  & 63.3  & 39.9  & 59.1  & 66.7  & 74.5  & 86.6 & 91.0  & 57.4  & 60.1  & 59.3  & 50.7  & \textbf{78.5} & 48.0  & 40.2  & 63.9  \\
    MoCo-MAID \cite{MoCo} & 83.3  & 66.1 & 40.3  & 55.3  & 66.9  & 74.2  & 85.3  & 90.8  & 60.5  & 58.3  & 57.2  & 52.7  & 77.2  & 46.7  & 38.5  & 63.6  \\
    MoCo-IN-MAID \cite{MoCo} & 83.4  & 62.1  & 39.9  & 53.8  & 66.3  & 74.5  & 85.3  & 91.8  & 59.0  & 58.6  & 54.8  & 56.6  & 76.9  & 48.9  & 33.8  & 63.0  \\
    MoCov2 \cite{MOCOv2} & 83.4  & 58.3  & 40.4  & 58.4  & 67.7  & 75.3  & 86.2  & 91.8  & 54.2  & 58.4  & 59.5  & 56.8  & 78.1  & 51.2  & 42.3  & 64.2  \\
    SeCo \cite{SeCo} & 82.0  & 57.3  & 35.8  & 42.6  & 65.4  & 76.3  & 86.5  & 90.8  & 39.2  & 57.7  & 51.8  & 46.7  & 75.2  & 50.5  & 34.7  & 59.5  \\
    Geo-Aware \cite{GeoAware} & 84.1  & 66.1  & 36.9  & 56.0  & \textbf{68.0} & 76.2  & 86.0  & 91.6  & 53.5  & 61.5  & 61.4 & 52.1  & 74.3  & 51.0  & 43.8  & 64.2  \\
    CACO \cite{CACO} & 82.0  & 63.6  & 38.3  & 49.1  & 67.5  & 76.9  & 85.5  & 91.2  & 44.4  & 57.8  & 55.8  & 52.8  & 75.0  & 46.6  & 37.0  & 61.6  \\
    TOV \cite{TOV}   & 83.7  & \textbf{66.5}  & 41.4  & 60.6  & 67.4  & 76.3  & \textbf{86.8}  & 91.6  & 60.1  & 61.5  & \textbf{62.0}  & 55.8  & 75.2  & 51.4  & 45.6  & 65.6  \\
    GeRSP & \textbf{85.3} & 63.6  & 42.7  & 60.7  & 67.2  & 75.2  & 85.9  & 91.7  & 60.4  & 63.0  & 61.2  & \textbf{56.8} & 78.2  & 50.4  & \textbf{46.5} & 65.9  \\
    GeRSP-200 & 85.2  & 64.3  & \textbf{44.9} & 60.5  & 65.4  & 75.0  & 86.1  & \textbf{92.8}  & \textbf{65.0} & \textbf{63.2} & 60.4  & 56.3  & 76.7  & 50.5  & 44.7  & \textbf{66.1} \\
    \midrule
    \textit{RetinaNet} &       &       &       &       &       &       &       &       &       &       &       &       &       &       &       &  \\
    From Scratch & 74.7  & 41.1  & 17.7  & 22.8  & 57.0  & 64.6  & 71.1  & 83.8  & 9.2   & 40.7  & 21.8  & 26.3  & 67.5  & 33.4  & 8.8   & 42.7  \\
    IMP   & 83.3  & 64.6  & 34.1  & 54.6  & 58.2  & 70.6  & 74.5  & 92.8  & 55.1  & 57.4  & 54.7  & 54.9  & 74.0  & 48.1  & 25.9  & 60.2  \\
    MoCo-IN \cite{MoCo} & 83.6  & 59.2  & 32.6  & 49.9  & 61.2 & 71.8  & 76.5  & 92.3  & 54.8  & \textbf{60.0} & 49.5  & 46.6  & 76.8  & 45.9  & 23.2  & 58.9  \\
    MoCo-MAID \cite{MoCo} & 84.3  & 62.3  & 35.6  & 52.0  & 60.8  & 72.1  & 76.5 & 92.0  & 57.3  & 58.6  & 47.5  & 50.1  & 77.3  & 47.1  & 24.4  & 59.9  \\
    MoCo-IN-MAID \cite{MoCo} & 82.8  & 62.8  & 34.5  & 45.5  & 58.4  & 71.9  & 75.4  & 93.0  & 53.8  & 57.2  & 48.6  & 51.6  & 75.7  & 46.5  & 29.9  & 59.2  \\
    MoCov2 \cite{MOCOv2} & 83.7  & 63.8  & 35.0  & 53.3  & 60.2  & 71.6  & 76.4  & 91.8  & 51.9  & 57.4  & 48.6  & 48.6  & 77.0  & 48.1  & 32.3  & 60.0  \\
    SeCo \cite{SeCo} & 80.8  & 56.9  & 30.1  & 38.7  & 60.8  & 69.1  & 75.1  & 89.5  & 41.9  & 53.9  & 43.2  & 43.2  & 72.8  & 43.7  & 16.6  & 54.4  \\
    Geo-Aware \cite{GeoAware} & 81.0  & 62.3  & 27.6  & 52.9  & 59.2  & 67.3  & 73.8  & 91.3  & 53.5  & 54.6  & 46.8  & 49.5  & 71.2  & 47.1  & 26.6  & 57.7  \\
    CACO \cite{CACO} & 79.9  & 60.0  & 32.7  & 42.3  & 59.0  & 68.3  & 74.9  & 91.2  & 47.4  & 53.1  & 43.4  & 50.9  & 74.0  & 43.5  & 22.5  & 56.2  \\
    TOV \cite{TOV} & 82.3  & \textbf{66.1}  & 36.9  & 51.9  & \textbf{61.4}  & 70.7  & \textbf{76.9}  & 91.0  & 52.6  & 57.8  & 50.8  & \textbf{57.3}  & 74.5  & 49.9  & 29.0  & 60.6  \\
    GeRSP & 84.9  & 65.6 & 36.3  & 54.3  & 60.0  & 71.3  & 74.7  & 93.7  & \textbf{67.4} & 59.2  & 54.8  & 56.4  & 75.8  & 50.1  & 31.8  & 62.4  \\
    GeRSP-200 & \textbf{85.5} & 65.3  & \textbf{38.1} & \textbf{55.5} & 60.7  & \textbf{72.5} & 75.9  & \textbf{94.2} & 65.2  & 58.7  & \textbf{54.9} & \textbf{57.3} & \textbf{77.4} & \textbf{50.6} & \textbf{32.6} & \textbf{63.0} \\
    \midrule
    \textit{DyHead} &       &       &       &       &       &       &       &       &       &       &       &       &       &       &       &  \\
    From Scratch & 79.7  & 49.2  & 27.1  & 37.1  & 63.0  & 72.1  & 81.9  & 86.9  & 26.6  & 52.1  & 23.3  & 37.9  & 75.4  & 46.1  & 18.5  & 51.8  \\
    IMP   & 85.5  & 64.4  & 40.3  & 59.7  & 66.6  & 76.1 & 84.9  & 92.6  & 54.0  & 65.5  & 47.9  & 54.7  & 79.9  & 49.1  & \textbf{51.0} & 64.8  \\
    MoCo-IN \cite{MoCo} & 85.4  & 59.6  & 37.9  & 52.7  & \textbf{68.7} & 75.5  & 84.1  & 92.9  & 59.7  & 66.6  & 48.7  & 51.9  & \textbf{81.2} & 48.1  & 35.4  & 63.2  \\
    MoCo-MAID \cite{MoCo} & 84.2  & 65.5  & 40.5  & 55.1  & 67.7  & 74.7  & 84.2  & 93.2  & 60.3  & 66.7  & 49.4  & 52.3  & 80.7  & 46.2  & 32.2  & 63.5  \\
    MoCo-IN-MAID \cite{MoCo} & 84.7  & 65.1  & 39.3  & 53.5  & 67.1  & 74.6  & 84.2  & 92.8  & 58.1  & 66.3  & 45.5  & 54.5  & 79.8  & 47.1  & 31.2  & 62.9  \\
    MoCov2 \cite{MOCOv2} & 85.2  & 62.3  & 40.2  & 49.5  & 67.0  & 74.9  & 84.1  & 92.6  & 53.7  & 66.4  & 46.8  & 51.8  & 80.5  & 50.0  & 38.6  & 62.9  \\
    SeCo \cite{SeCo}  & 82.1  & 61.0  & 35.0  & 46.5  & 63.8  & 72.3  & 83.2  & 89.5  & 44.9  & 59.4  & 31.1  & 46.1  & 78.7  & 47.6  & 28.4  & 58.0  \\
    Geo-Aware \cite{GeoAware} & 83.8  & 62.7  & 37.4  & 57.3  & 65.0  & 74.4  & 83.6  & 90.7  & 48.0  & 64.5  & 45.3  & 58.8  & 78.5  & 48.9  & 37.6  & 62.4  \\
    CACO \cite{CACO} & 82.6  & 59.7  & 38.7  & 56.2  & 65.0  & 75.9  & 83.7  & 90.2  & 48.2  & 61.1  & 40.9  & 51.8  & 79.3  & 44.8  & 28.8  & 60.5  \\
    TOV \cite{TOV}   & 85.5  & 66.0  & 44.1  & 59.3  & 64.8  & \textbf{76.2}  & 84.4  & 91.4  & 61.8  & 63.6  & 46.9  & 56.5  & 80.2  & 49.5  & 42.8  & 64.9  \\
    GeRSP & 87.0  & 66.3  & 42.4  & 60.3  & 68.3  & 75.7  & \textbf{85.0} & \textbf{93.7} & \textbf{65.5} & 68.5  & 51.5  & 56.5  & 80.0  & 51.2  & 40.2  & 66.1  \\
    GeRSP-200 & \textbf{87.3} & \textbf{66.9} & \textbf{44.3} & \textbf{62.4} & 67.7  & 76.1 & 84.5  & 93.3  & 65.3  & \textbf{69.5} & \textbf{52.5} & \textbf{60.5} & 80.5  & \textbf{51.3} & 49.0  & \textbf{67.4} \\
    \bottomrule
    \end{tabular}%
  \label{tab:DOTA}%
\end{table*}%

\textbf{Object Detection:} 
Table \ref{tab:DIOR} compares the detection performance of GeRSP on DIOR with other pre-training methods. 
The comparison is made by fine-tuning with Faster R-CNN \cite{Faster-RCNN}, RetinaNet \cite{RetinaNet}, and DyHead \cite{Dy-Head}. 
We observe that GeRSP consistently outperforms other pre-training methods. 
Compared to IMP, GeRSP improves mAP from 65.1\% to 67.1\% (+2.0\%) on Faster R-CNN, improves mAP from 66.5\% to 67.5\% (+1.0\%) on RetinaNet, improves 70.6\% to 71.2\% (+0.6\%) on DyHead. 
MoCo-MAID only uses MAID for training, so it can only obtain domain knowledge about remote sensing images and lacks more robust generalization performance. 
Compared to MoCo-MAID, GeRSP improves mAP from 66.1\% to 67.1\% (+1.0\%) on Faster R-CNN, GeRSP improves mAP from 64.2\% to 67.5\% (+3.3\%) on RetinaNet, GeRSP improves 68.5\% to 71.2\% (+2.7\%) on DyHead. 

As evidenced by Table \ref{tab:DIOR}, MoCo-IN-MAID, despite being pre-trained using both the ImageNet and MAID datasets, fails to enhance performance on downstream tasks and exhibits a certain degree of degradation. 
Conversely, our GeRSP framework consistently achieves effective performance improvement. 
This observation underscores GeRSP's ability to effectively leverage information from multiple datasets, resulting in the acquisition of more generalized features.
Additionally, it is worth noting that the performance of pre-trained weights obtained through existing pre-training methods in the detection task displays instability, with their fine-tuning performance on the three detection methods slightly trailing behind that of IMP. 
In contrast, GeRSP exhibits exceptional stability in its performance in detection while continuously demonstrating improvement over time. 

Table \ref{tab:DOTA} compares the fine-tuning effects of different pre-training methods on DOTA. 
We employed Faster R-CNN, RetinaNet, and DyHead as the detection models to validate the results. 
GeRSP consistently achieves notable improvements in detector performance and outperforms other methods across most regions. 
Specifically, GeRSP proves more advantageous than IMP in detection tasks, showcasing an increase of +0.8\% on Faster R-CNN, +2.2\% on RetinaNet, and +1.3\% on DyHead. 
Furthermore, GeRSP demonstrates more significant improvement than MoCo-MAID, with gains of +2.3\% on Faster R-CNN, +2.5\% on RetinaNet, and +2.6\% on DyHead. 

Notably, MoCo-MAID, which solely utilizes remote sensing images, fails to exhibit any advantages over IMP in the detection task, showing decreases of -1.5\% on Faster R-CNN, -0.3\% on RetinaNet, and -1.3\% on DyHead.
Moreover, the more advanced self-supervised pre-training method MoCov2 fails to yield performance improvements. 
In contrast, our GeRSP method effectively enhances detection performance. 
Moreover, from the GeRSP-200 row, it is evident that extended training further enhances detection performance. GeRSP proves highly effective in elevating the capabilities for detection.

\textbf{Segmentation:} 
Table \ref{tab:segmentation} compares mIoU of PSANet \cite{PSANet} and DeepLabV3+ \cite{DeepLabV3+} initialized by GeRSP and other pre-training methods on the LoveDA dataset. 

For PSANet, GeRSP consistently outperforms other pre-training methods on LoveDA. 
Compared with MoCo-IN, MoCo-MAID, and MoCo-IN-MAID, GeRSP improves mIoU by +0.99\%, +2.38\%, and +1.19\%, which demonstrates that both pre-training dataset and method play essential roles in representation learning. 
Compared to GeRSP, MoCo-IN exhibits poor performance due to its insufficient understanding of knowledge in the RS domain.
Although MoCo-MAID utilizes remote sensing images for pre-training and acquires domain knowledge specific to such images, it lacks generalization ability. 
In the case of MoCo-IN-MAID, the pre-training method also influences the pre-training process. 

For DeepLabV3+, GeRSP outperforms all other supervised pre-training and self-supervised pre-training methods on LoveDA. 
As shown in Table \ref{tab:segmentation}, GeRSP improves mIoU by +2.42\%, +1.86\%, and +0.8\% when compared with MoCo-IN, MoCo-MAID, and MoCo-IN-MAID. 
The results are the same as in PSANet, demonstrating that GeRSP efficiently learns representations for remote sensing downstream tasks. 
Compared with SeCo, GeRSP consistently improves mIoU on the two semantic methods, especially +4.89\% for PSANet and +5.43\% for DeepLabV3+. 
This observation shows that GeRSP can effectively enhance semantic segmentation performance.

\begin{table}[htbp]
  \centering
    \setlength\tabcolsep{3pt}
  \caption{Results of PSANet \cite{PSANet} and DeepLabV3+ \cite{DeepLabV3+} with different pre-trained backbones on LoveDA \cite{LoveDA} dataset.}
    \begin{tabular}{c|c|cc}
    \toprule
    \multirow{2}[2]{*}{Pre-training} & \multirow{2}[2]{*}{Dataset} & PSANet & DeepLabv3+ \\
          &       & mIoU  & mIoU \\
    \midrule
    IMP   & ImageNet & 49.24 & 48.39 \\
    MoCo-IN \cite{MoCo} & ImageNet & 48.54 & 46.64 \\
    MoCo-MAID \cite{MoCo} & MAID  & 47.15 & 47.20 \\
    MoCo-IN-MAID \cite{MoCo} & ImageNet + MAID & 48.34 & 48.26 \\
    MoCov2 \cite{MOCOv2} & MAID  & 44.79 & 48.57 \\
    SeCo \cite{SeCo} & 1M Sentinel-2 & 44.64 & 43.63 \\
    Geo-Aware \cite{GeoAware} & GeoImageNet & 49.37 & 48.76 \\
    CACO \cite{CACO} & CACO 1M &   48.81    &  48.89\\
    TOV \cite{TOV} & TOV-NI + TOV-RS &  49.33  &  49.7  \\
    \textbf{GeRSP} & ImageNet + MAID & 49.53 & 49.06 \\
    \textbf{GeRSP-200} & ImageNet + MAID & \textbf{49.56} & \textbf{50.56} \\
    \bottomrule
    \end{tabular}%
  \label{tab:segmentation}%
\end{table}%

\begin{table}[htbp]
  \centering
    \setlength\tabcolsep{2pt}
  \caption{Sensitivity of balance coefficient $\alpha$ in GeRSP.}    \begin{tabular}{c|c|c|c|c|c|c}
    \toprule
    \multirow{2}[2]{*}{$\alpha$} & EuroSAT & NWPU-RESISC45 & DOTA  & DIOR  & PSANet & DeepLabv3+ \\
          & Top-1 & Top-1 & mAP   & mAP   & mIoU  & mIoU \\
    \midrule
    0 & 97.52 ± 0.10 & 91.19 ± 0.15 & 63.6  & 66.1  & 47.15 & 47.20 \\
    0.5 & 97.79 ± 0.10 & 92.54 ± 0.09 & 65.6  & 67.6  & 49.37 & 49.36 \\
    1 & 97.87 ± 0.15 & 92.67 ± 0.16 & 65.9  & 67.1  & 49.53 & 49.06 \\
    \bottomrule
    \end{tabular}%
  \label{tab:alpah}%
\end{table}%

\begin{figure*}[!htb]
	\centering
	\includegraphics[width=1\linewidth]{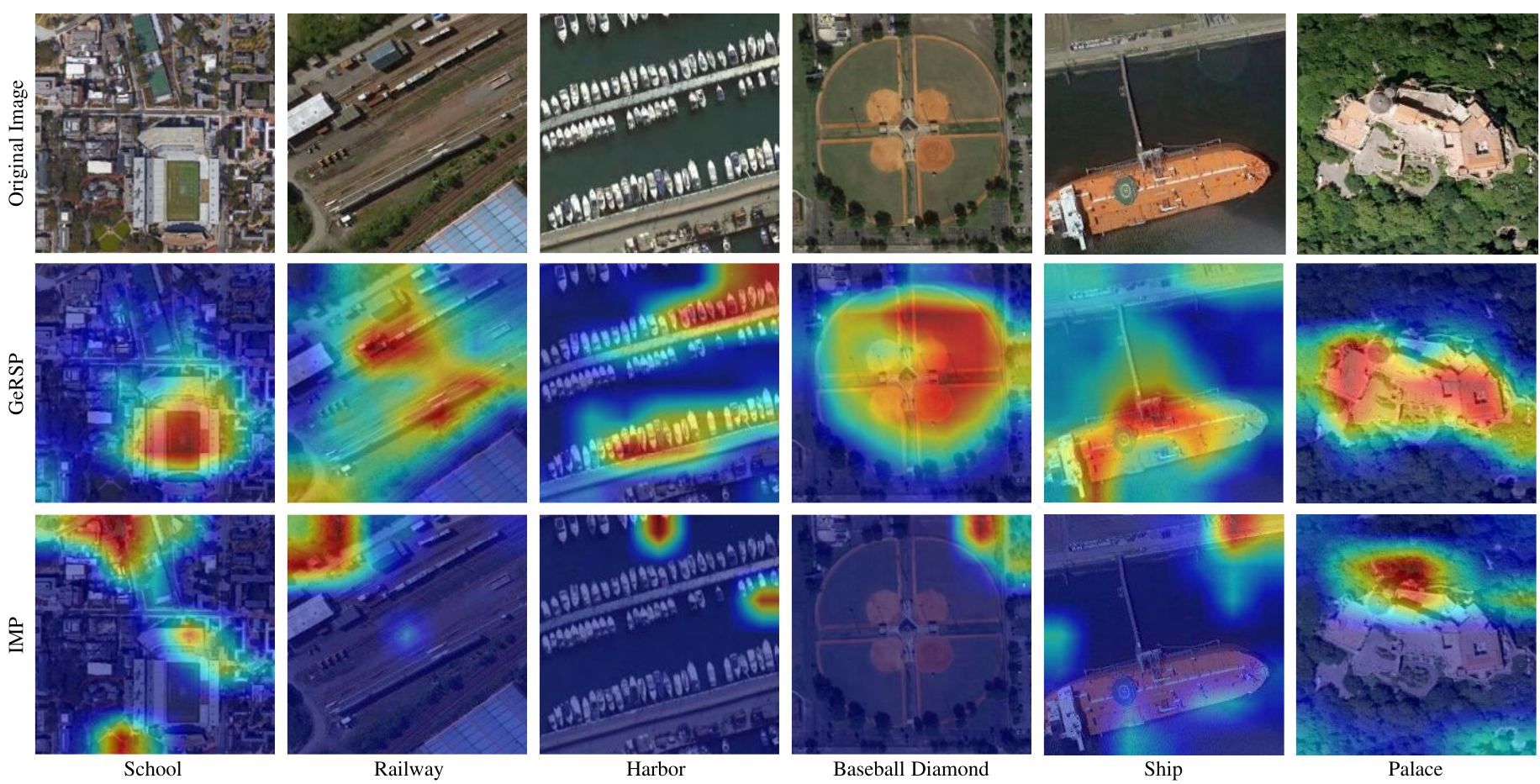}
	\caption{Class activation maps (CAMs) visualization of GeRSP model and IMP model on six categories.}
	\label{fig:CAM1}
\end{figure*}

\begin{figure}[!htb]
    \centering
    \includegraphics[width=\linewidth]{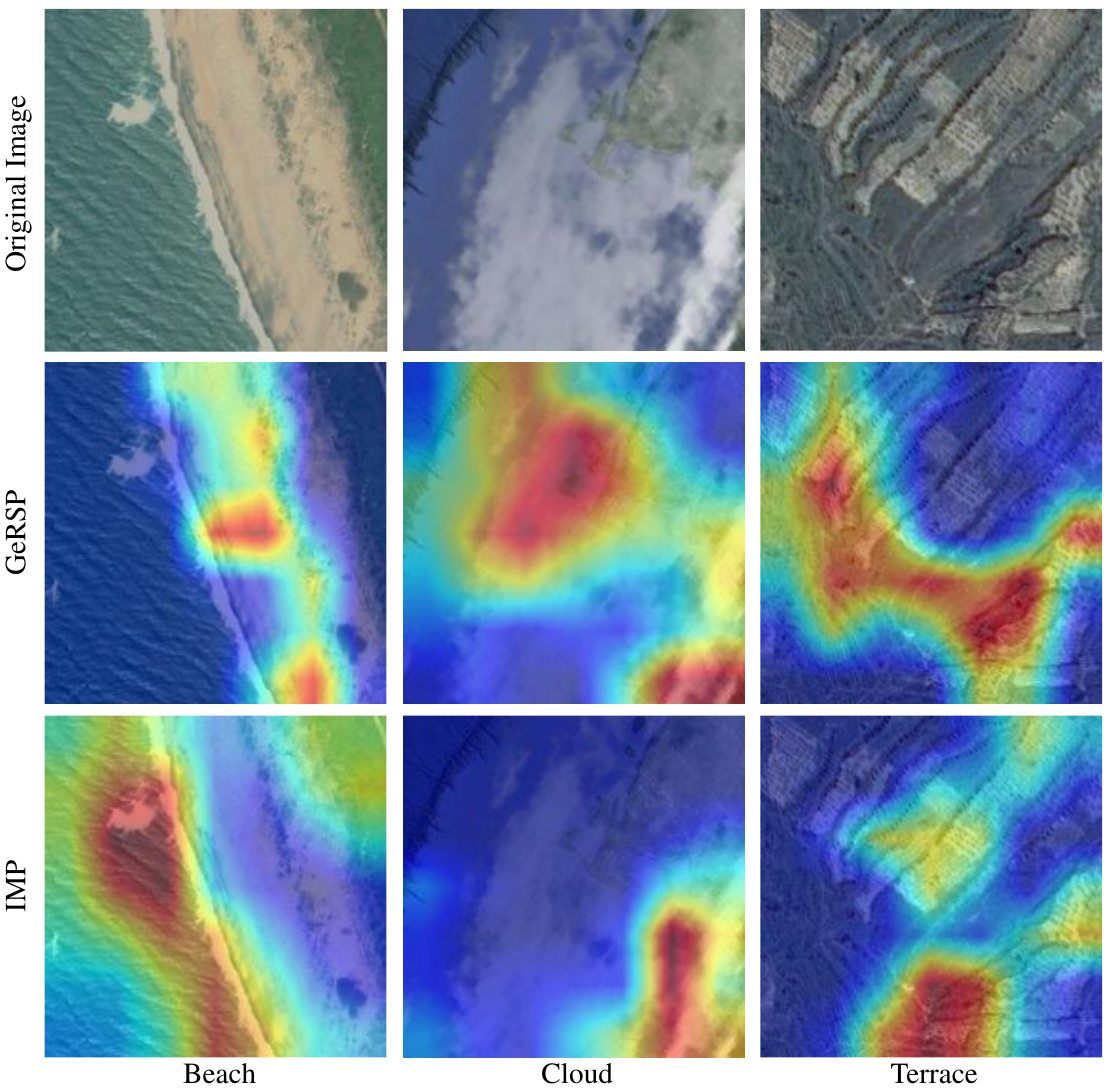}
	\caption{Class activation maps (CAMs) visualization of GeRSP model and IMP model on beach, cloud, and terrace.}
	\label{fig:CAM2}
\end{figure}

\begin{figure}[!htb]
    \centering
    \includegraphics[width=\linewidth]{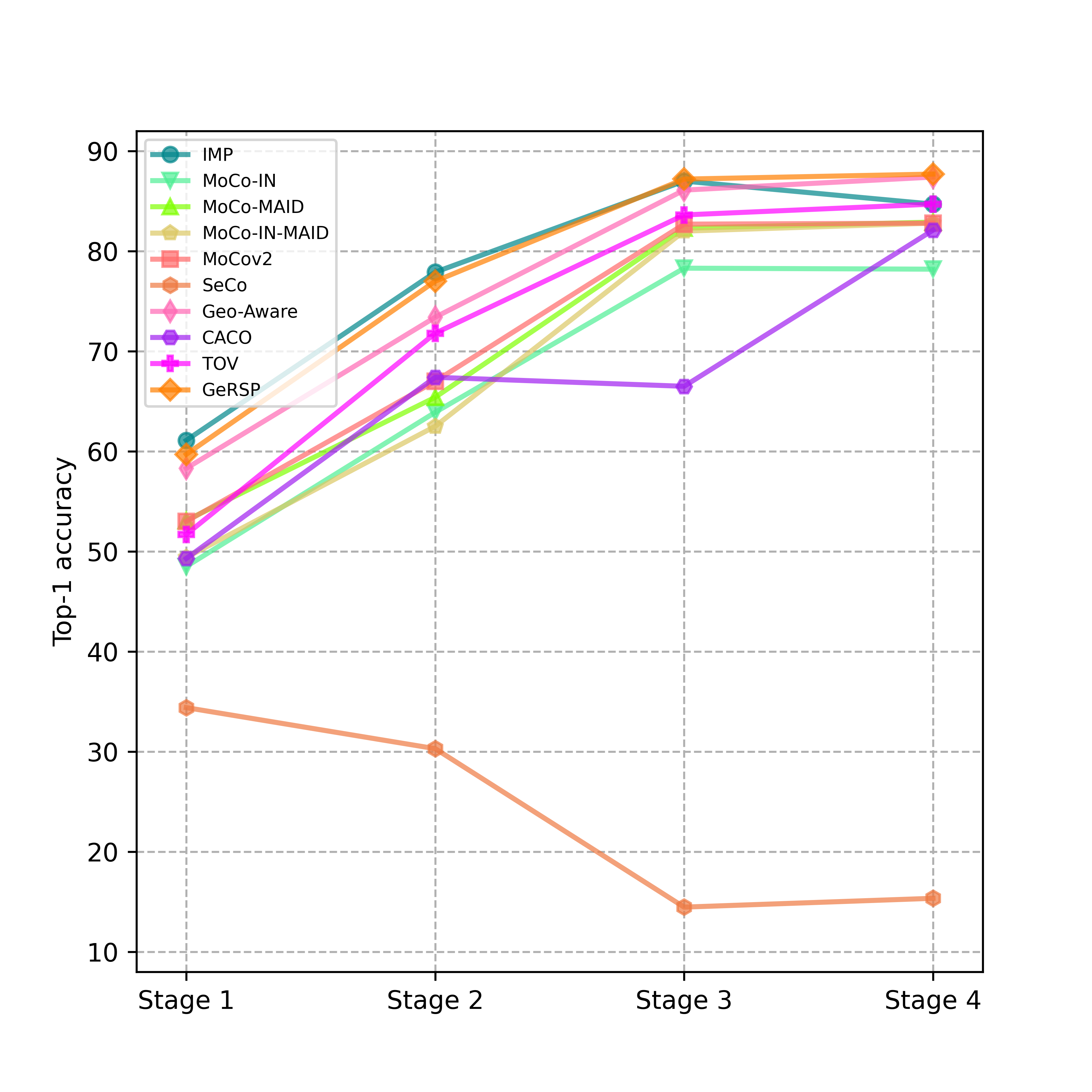}
	\caption{Top-1 accuracy of stage-wise evaluation \cite{wang2022revisiting} on NWPU-RESISC45 \cite{NWPU_RESISC45}. All parameters of the pre-trained models are frozen during the training process. During training, features from various stages are extracted, flattened, and fed into a linear layer for classification, achieving a stage-wise linear evaluation.}
	\label{fig:LinearProb}
\end{figure}

\subsection{Visualization}
Fig. \ref{fig:CAM1} visually explains model predictions on scene classification tasks through class activation maps (CAM). 
The CAM is implemented by Grad-CAM++ \cite{GradCam++}. 
The principle of Grad-CAM++ \cite{GradCam++} is to use a weighted combination of the positive partial derivatives as weights to sum activation maps and construct CAM capable of explaining numerous instances. 
We compare the models obtained by INT Supervised and GeRSP, as shown in the second and third rows of Fig. \ref{fig:CAM1}. 
Intuitively, the GeRSP model can focus well on the instances in RS images, such as school, palace, and baseball diamond. 
It can also sense multiple objects in the image, such as ships in the harbor shown in the third column of Fig. \ref{fig:CAM1}.  
This demonstrates that GeRSP can improve the model's ability to extract semantic information from RS images. 
On the contrary, the IMP model is difficult to capture RS semantic information and lacks the generalization ability in RS cognitive tasks. 
Fig. \ref{fig:CAM2} shows that the IMP model struggles to recognize the scene categories in images, whereas the GeRSP model can recognize these scene concepts through unsupervised learning. 

We also undertake a stage-wise linear evaluation, as outlined by Wang et al. \cite{wang2022revisiting}, to elucidate the underlying sources of generalization of GeRSP. 
Following \cite{wang2022revisiting}, we freeze the parameters of the pre-trained model and subsequently conduct linear evaluation training using features obtained from each stage. 
The results of linear evaluation reflect the semantic information contained in the features.  
Additionally, since the quality of low-mid level features significantly impacts the performance of semantic discrimination tasks, the performance of linear evaluation indirectly reflects the caliber of low-mid level features. 
As shown in Fig. \ref{fig:LinearProb}, overall, as we progress through the stages, there is a gradual enhancement of semantic information, resulting in a continuous improvement in linear evaluation performance. 
Compared with other competitive methods, it is evident that both IMP and our GeRSP demonstrate comparable linear evaluation performance across different stages, consistently outperforming other methods at every stage, especially in the first and second stages. 
This phenomenon suggests that ImageNet supervised training can acquire high-quality image features across different levels, particularly emphasizing low to mid-level features, which is advantageous for semantic discrimination tasks in remote sensing images. 
Therefore, our GeRSP, through the introduction of natural image supervised learning, acquires rich generic image knowledge that can be effectively applied to subsequent remote sensing tasks. 

We conduct experiments on the balance coefficient $\alpha$ in Eq. \ref{eq:TotalLoss}. 
As shown in Table \ref{tab:alpah}, when there is no supervised loss, e.g., $\alpha$ set to 0, the model's performance decreases across all tasks. 
When $\alpha$ is set to 0.5 or 1, the natural image supervised loss is introduced into the pre-training, significantly improving performance. 
Also, from the results, the scale of $\alpha$ has a relatively minor impact on the pre-training effectiveness, demonstrating the insensitivity of GeRSP to $\alpha$.


\section{Conclusion}
This paper proposes Generic Knowledge Boosted Remote Sensing Pre-training (GeRSP) for remote sensing pre-training. 
GeRSP aims to obtain a pre-trained model suitable for initializing deep learning models for RS understanding tasks. 
The proposed method leverages a teacher-student architecture to harness the benefits of both supervised pre-training and self-supervised pre-training, mitigating the impact of domain gaps between RS images and natural images. 
During the self-supervised pre-training process, GeRSP acquires domain-specific features from unlabeled RS images. In contrast, in the supervised pre-training process, it learns general knowledge from labeled natural images. 
By integrating self-supervised and supervised pre-training, GeRSP simultaneously learns representations with general and special knowledge. 
Subsequently, GeRSP's effectiveness is evaluated by conducting three remote sensing downstream tasks: object detection, semantic segmentation, and scene classification. 
Consistently, the experimental results demonstrate that GeRSP is an effective pre-training method in remote sensing, enhancing the performance of various downstream tasks.

GeRSP effectively mitigates the limitations of contrastive learning in perceiving fine-grained features in RS images, ensuring transferability across various tasks, especially segmentation and detection tasks. 
In addition to introducing ImageNet supervised learning, using masked image modeling \cite{MAE, CMID, I-JEPA, wang2023feature, li2023self, wei2022masked} and explicitly specified learning of image descriptors \cite{wang2023feature, li2023self, wei2022masked} can achieve similar goals. 
Replace supervised training with these methods, potentially resulting in enhanced generalization performance, which we will further consider in future. 



\ifCLASSOPTIONcaptionsoff
  \newpage
\fi
\bibliographystyle{IEEEtran}
\bibliography{paper}
\begin{IEEEbiography}
[{\includegraphics[width=1in,height=1.25in,clip,keepaspectratio]{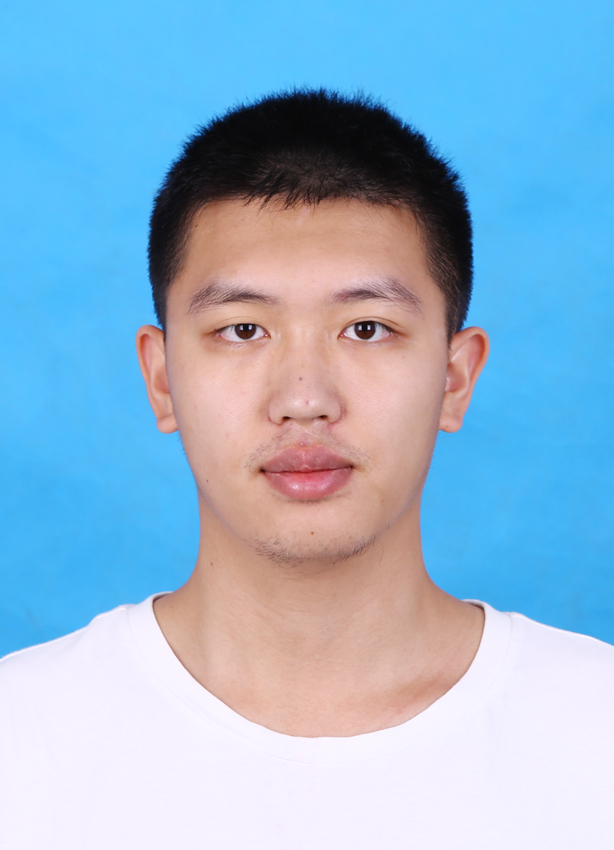}}]
{Ziyue Huang}
received the BS degree in information and computing science from the School of Mathematics and Statistics, Beijing Institute of Technology, Beijing, China, in 2018. He is currently pursuing the Ph.D. degree in computer science with the Laboratory of Intelligent Recognition and Image Processing, School of Computer Science and Engineering, Beihang University. His research interests include computer vision and object detection.
\end{IEEEbiography}

\begin{IEEEbiography}
[{\includegraphics[width=1in,height=1.25in,clip,keepaspectratio]{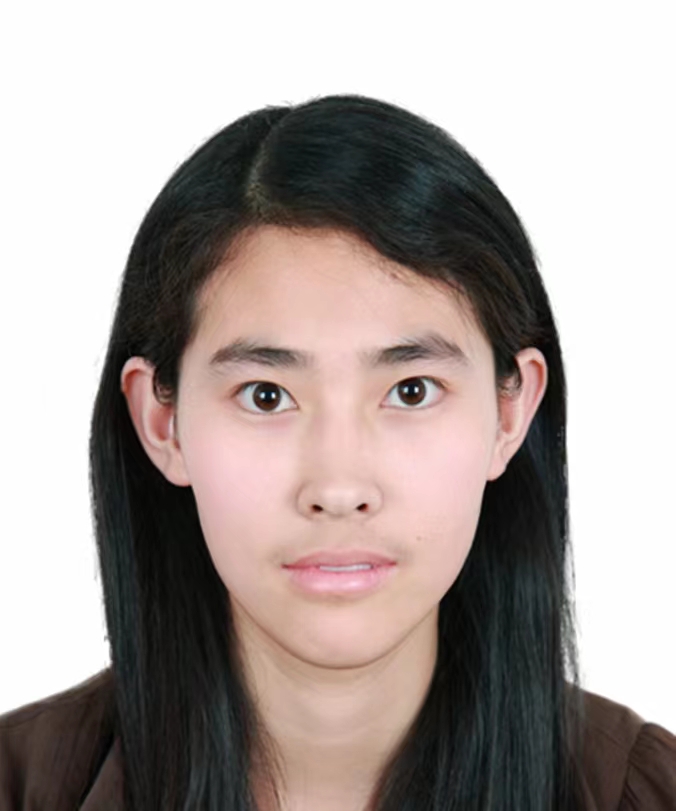}}]
{Mingming Zhang} 
received the BS degree in information and computer science from Liaoning University, Shenyang, China and the MS degree in software engineering from Beihang University, Beijing, China. She is currently pursuing the Ph.D. degree in computer science with the Laboratory of Intelligent Recognition and Image Processing, School of Computer Science and Engineering, Beihang University. Her research interests include remote sensing image analysis and computer vision.
\end{IEEEbiography}

\begin{IEEEbiography}
[{\includegraphics[width=1in,height=1.25in,clip,keepaspectratio]{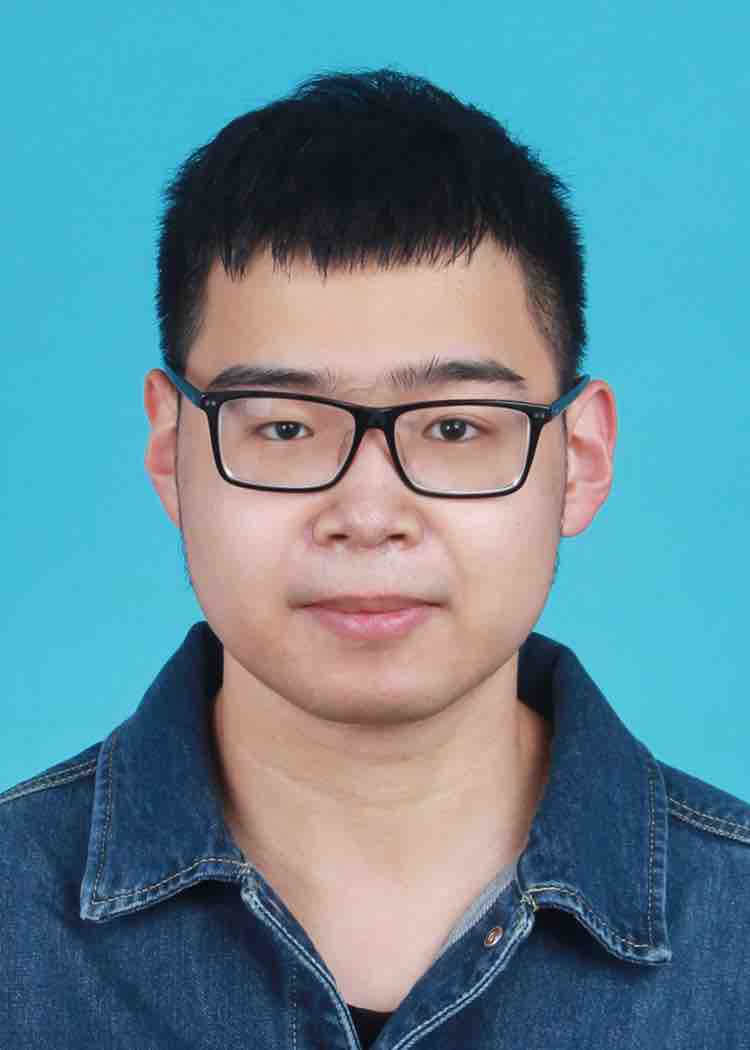}}]
{Yuan Gong}
received the BS degree in computer science from the School of Computer Science and Engineering at Beihang University, Beijing, China, in 2022. He is currently pursuing a Master’s degree in software engineering at the School of Software and Microelectronics, Peking University, Beijing, China. 
\end{IEEEbiography}

\begin{IEEEbiography}
[{\includegraphics[width=1in,height=1.25in,clip,keepaspectratio]{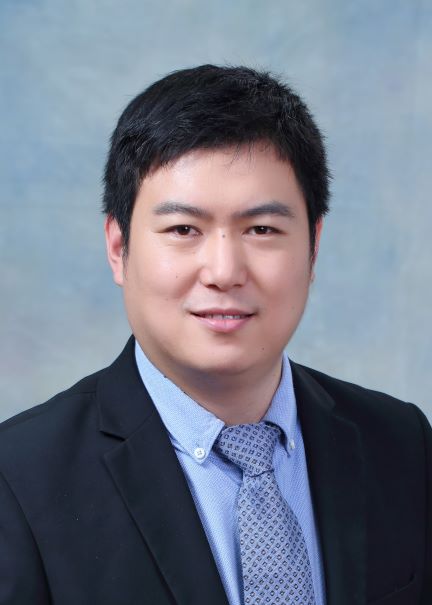}}]
{Qingjie Liu}
received the BS degree in computer science from Hunan University, Changsha, China, in 2007, and the Ph.D. degree in computer science from Beihang University, Beijing, China, in 2014. He is currently an Associate Professor with the School of Computer Science and Engineering, Beihang University. He is also a Distinguished Research Fellow with the Hangzhou Institute of Innovation, Beihang University, Hangzhou. His current research interests include image fusion, object detection, image segmentation, and change detection. He is a member of the IEEE.
\end{IEEEbiography}

\begin{IEEEbiography}
[{\includegraphics[width=1in,height=1.25in,clip,keepaspectratio]{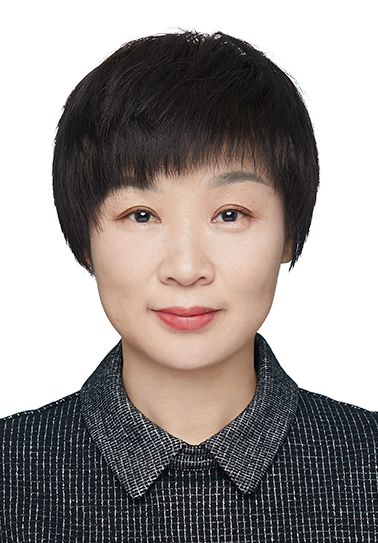}}]
{Yunhong Wang}
received the BS degree in electronic engineering from Northwestern Polytechnical University, Xi’an, China, in 1989, and the MS and Ph.D. degrees in electronic engineering from the Nanjing University of Science and Technology, Nanjing, China, in 1995 and 1998, respectively. 

She was with the National Laboratory of Pattern Recognition, Institute of Automation, Chinese Academy of Sciences, Beijing, China, from 1998 to 2004. Since 2004, she has been a Professor with the School of Computer Science and Engineering, Beihang University, Beijing, where she is also the Director of the Laboratory of Intelligent Recognition and Image Processing. Her research interests include biometrics, pattern recognition, computer vision, data fusion, and image processing. She is a Fellow of IEEE, IAPR, and CCF.
\end{IEEEbiography}

\end{document}